\documentclass{article} 
\PassOptionsToPackage{numbers,sort&compress}{natbib}
\usepackage{conference,times}
\setcitestyle{numbers,square,comma}

\usepackage{amsmath,amsfonts,bm}
\usepackage{amssymb}
\usepackage{array}
\usepackage{multirow}
\usepackage{graphicx}
\usepackage{rotating}
\usepackage{placeins}
\usepackage{booktabs}
\usepackage{adjustbox}

\AddToHook{env/table/begin}{\setlength{\belowcaptionskip}{11pt}}
\AddToHook{env/table*/begin}{\setlength{\belowcaptionskip}{11pt}}
\AddToHook{env/sidewaystable/begin}{\setlength{\belowcaptionskip}{11pt}}
\AddToHook{env/sidewaystable*/begin}{\setlength{\belowcaptionskip}{11pt}}
\usepackage{enumitem}
\usepackage{xcolor}
\usepackage{tikz}
\usetikzlibrary{positioning,arrows.meta,calc,fit,backgrounds,shapes.geometric,matrix,decorations.pathreplacing}
\usepackage{algorithm}
\usepackage{algpseudocode}
\usepackage{amsthm}

\usepackage{hyperref}
\usepackage{url}

\theoremstyle{definition}

\theoremstyle{plain}

\title{FlashCart: Fast Cartesian Tensor Products for Equivariant Interatomic Potentials}

\author{
\parbox{\dimexpr\textwidth-2\tabcolsep\relax}{%
\centering\normalfont
\textbf{Viktor Zaverkin}$^{1,2,3,\ast}$ \hspace{10pt}
\textbf{Payman Goodarzi}$^{1}$ \hspace{10pt}
\textbf{Sergey V. Sukhomlinov}$^{2}$\\
\textbf{Davit Hovhannisyan}$^{3}$ \hspace{9pt}
\textbf{Roland Aydin}$^{2,3}$ \hspace{9pt}
\textbf{Martin H. Müser}$^{2}$ \hspace{9pt}
\textbf{Mathias Niepert}$^{4,5}$\\
$^1$INM -- Leibniz Institute for New Materials
\quad $^2$Saarland University\\
$^3$German Research Center for Artificial Intelligence (DFKI)\\
$^4$University of Stuttgart \quad $^5$NEC Laboratories Europe
}}

\makeatletter
\newcommand{\ftype@preprintnotice}{8}
\newcommand{\preprintnotice}{%
  \@float{preprintnotice}[b]%
    \footnotesize Preprint. Under review.%
  \end@float
}
\makeatother

\iclrfinalcopy
\begin{document}

\maketitle
\lhead{}
\begingroup
\renewcommand{\thefootnote}{}
\footnotetext{$^\ast$\,Corresponding author:
  \href{mailto:viktor.zaverkin@leibniz-inm.de}
       {viktor.zaverkin@leibniz-inm.de}}
\endgroup
\preprintnotice

\begin{abstract}
Machine-learned interatomic potentials extend atomistic simulations beyond the length- and timescales accessible to electronic-structure methods. However, the computational cost of equivariant architectures limits the local correlations they can represent in practice and therefore their achievable accuracy. Here we introduce FlashCart, which makes higher-order correlations affordable by combining generated GPU kernels with an architecture that recursively builds equivariant features and compresses them to a fixed width at each step. We express tensor products in independent Cartesian components and symbolically simplify them and their derivatives, producing fused kernels that often outperform optimized spherical counterparts. We then show that increasing correlation order improves accuracy more efficiently than increasing width, depth, or tensor rank. On SPICE-MACE-OFF, FlashCart models advance the measured accuracy--efficiency frontier: a model with $5.6$ million parameters achieves lower energy and force errors and $10\times$ faster inference than a transformer with $189$ million parameters.
\end{abstract}

\section{Introduction}
\label{sec:intro}


Machine-learned interatomic potentials (MLIPs) enable molecular and materials simulations at length- and timescales inaccessible to electronic-structure methods, but their practical application depends on the cost of repeated energy and force evaluations. Equivariant models such as NequIP~\citep{Batzner2022} and MACE~\citep{Batatia2022} describe each atom's neighborhood using features that transform under rotations as scalars, vectors, or higher-rank tensors and are coupled through equivariant products~\citep{Musaelian2023,Liao2023,Bochkarev2024,Lysogorskiy2026,Li2026,Zaverkin2024,Chen2026}. These products can also be applied repeatedly to capture increasingly complex relationships among neighboring atoms, building higher-order correlations as in the atomic cluster expansion~\citep{Drautz2019}. Higher orders can improve accuracy but increase evaluation cost. Methods like the Point Edge Transformer (PET) instead relax explicit rotational constraints~\citep{Pozdnyakov2023}, with recent models achieving state-of-the-art accuracy--efficiency tradeoffs~\citep{Mazitov2025b,Bigi2026}. Making higher-order correlations affordable offers a complementary route to improved accuracy--efficiency tradeoffs while retaining explicit equivariance.

Higher-order correlations can be built by successively combining the same input features in different ways, called coupling paths. Retaining all these paths increases the number of intermediate features, so compressing them between successive correlation orders controls this growth~\citep{Nigam2020}. The cost of individual products can also be reduced by fusing operations in GPU kernels to limit intermediate storage and memory traffic, as shown for spherical tensors~\citep{Geiger2024,Bharadwaj2025,Tan2026,Chorosajev2026}. Cartesian approaches provide an alternative to spherical tensor representations~\citep{Zaverkin2024,Chen2026}, with Ref.~\citenum{Chen2026} identifying dedicated GPU kernels as a promising direction. Such kernels can use the $2l+1$ independent components of a rank-$l$ irreducible Cartesian tensor, matching the dimension of its spherical counterpart~\citep{Applequist1989}. Writing the products in these components gives polynomial expressions that can be simplified and differentiated symbolically before compilation. Further details on related work are provided in Appendix~\ref{app:related-details}.

\begin{figure}[tbp]
\centering
\includegraphics[width=0.95\textwidth, trim=0 6pt 0 9pt, clip]{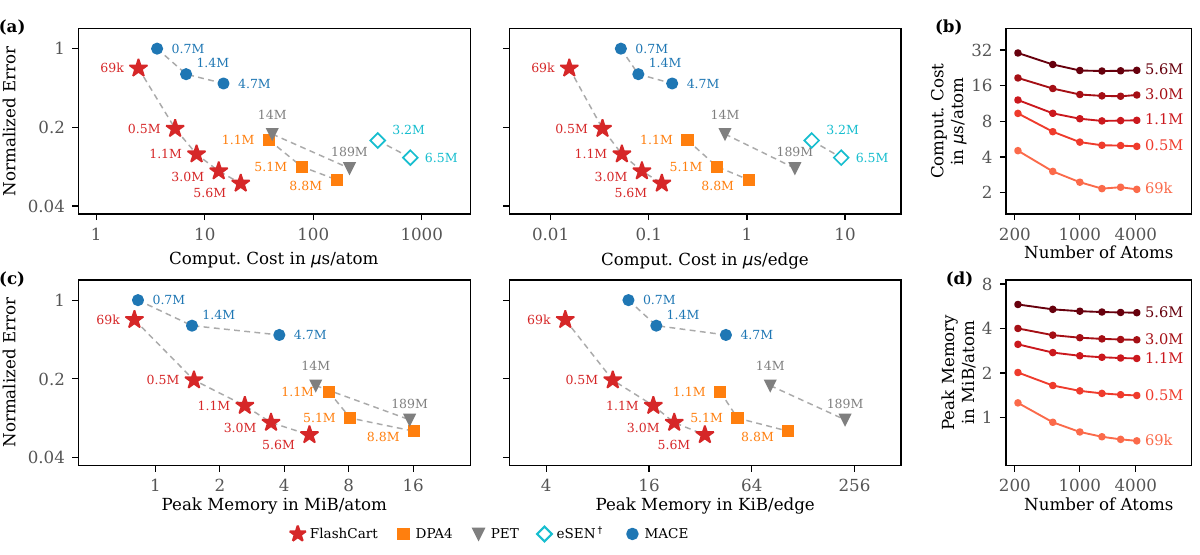}
\caption[Accuracy--efficiency tradeoff evaluated on SPICE-MACE-OFF.]{\textbf{Accuracy--efficiency tradeoff evaluated on SPICE-MACE-OFF.} (a,c) Normalized error as a function of evaluation time and peak GPU memory, with FlashCart models on the measured accuracy--efficiency frontier. Local evaluation costs in (a,c) use a $1000$-atom diamond cell. (b,d) Evaluation time and peak GPU memory per atom as a function of system size, reaching a plateau at smaller system sizes for larger FlashCart models. \textsuperscript{$\dagger$}All timings use an NVIDIA RTX PRO 6000 Blackwell (96\,GB), except for eSEN literature values measured on an NVIDIA A100 (80\,GB)~\citep{Fu2025}.}
\label{fig:pareto}
\end{figure}

We introduce FlashCart, a symbolic Cartesian tensor framework that combines generated GPU kernels with a recursive equivariant architecture. Specifically, (i) we express tensor products and their derivatives in independent Cartesian components and automatically generate fused GPU kernels for node-local products and neighbor aggregation (Sections~\ref{sec:lowering} and~\ref{sec:schedule}). These kernels often outperform optimized spherical counterparts in our benchmarks. (ii) We build higher-order correlations through repeated products, mixing paths and channels back to fixed width at each step (Section~\ref{sec:arch}). Each additional order requires one product and one linear map, so the correlation cost grows linearly at fixed width and maximum tensor rank. (iii) On SPICE-MACE-OFF~\citep{Eastman2023,Kovacs2025}, we show that increasing correlation order improves accuracy more efficiently than increasing width, depth, or tensor rank. The resulting FlashCart models lie on the measured accuracy--efficiency frontier (Figure~\ref{fig:pareto}), while results on MAD~\citep{Mazitov2025} support their use as competitive general-purpose interatomic potentials.

\section{Background}
\label{sec:background}

\textbf{Machine-learned interatomic potentials.} MLIPs approximate electronic-structure solutions by a parametric map from an atomic configuration $\mathcal S=\{\bm r_u,Z_u\}_{u=1}^{N_{\mathrm{at}}}$, defined by atomic positions and species, to its potential energy. Under periodic boundary conditions, it also includes a cell, and local neighborhoods include periodic images obtained by translating atoms by integer combinations of the cell vectors. The model writes the energy as a sum of atomic contributions, $E=\sum_u E_u$, built from finite-cutoff neighborhoods. In the absence of external fields, $E$ is invariant under translations, global rotations and reflections $\bm R\in O(3)$ acting on the positions and cell, and permutations of atoms of the same species. Forces are obtained as $\bm F_u=-\nabla_{\bm r_u}E$ and transform as $\bm F_u\mapsto\bm R\bm F_u$. Deriving them from a scalar energy ensures a conservative force field and energy-conserving dynamics. MLIPs are usually trained on reference energies and forces, which requires differentiating the forces with respect to model parameters. Each operation in the energy computation must therefore support a vector--Jacobian product (VJP) for force evaluation that can itself be differentiated during training. We refer to these two derivative operations as the VJP and the VJP-of-VJP.

\textbf{Equivariant convolution and many-body correlations.} Architectures that build many-body correlations combine equivariant convolutions, which collect information from neighboring atoms, with repeated node-local products of the aggregated features~\citep{Batatia2022,Bochkarev2024,Zaverkin2024}. For each atom $u$, $\mathcal N(u)$ contains all atoms and periodic images within the cutoff $r_{\max}$, excluding the central atom in the reference cell. The edge vector $\bm r_{uv}$ points from atom $u$ to its neighbor $v$, including any periodic translation, with length $r_{uv}$ and unit direction $\hat{\bm r}_{uv}=\bm r_{uv}/r_{uv}$. We write $\bm T_l(\hat{\bm r}_{uv})$ for the rank-$l$ direction tensor of an edge and $\bm h_{vl}^{(t)}$ for the $d$ feature channels at $v$. The edge directions, product intermediates, and propagated node states can retain different maximum ranks, denoted $l_{\max}^{\mathrm{edge}}$, $l_{\max}^{\mathrm{prod}}$, and $l_{\max}^{\mathrm{prop}}$.

The convolution couples $\bm h_{vl_1}^{(t)}$ with $\bm T_{l_2}(\hat{\bm r}_{uv})$ along admissible paths $(l_1,l_2)\to l$, applies distance-dependent weights $\bm W_{l_1l_2l}^{(t)}(r_{uv})$, and sums over $v\in\mathcal N(u)$ to form the atomic basis $\bm A_u=\bigoplus_l\bm A_{ul}$. The node-local products are evaluated recursively as $\bm\phi_u^{(0)}=1$ and $\bm\phi_u^{(q)}=\mathcal W^{(q)}\Pi_{\le l_q}(\bm\phi_u^{(q-1)}\hat{\otimes}\bm A_u)$ for $q=1,\ldots,\nu$, where the projection $\Pi_{\le l_q}$ discards ranks above $l_q$, the linear maps $\mathcal W^{(q)}$ are equivariant, and every step contributes to the layer output. The state $\bm\phi_u^{(q)}$ has polynomial degree $q$ in $\bm A_u$. If each neighbor contribution to $\bm A_u$ depends only on the central atom and that neighbor, these products have body order at most $q+1$, including the central atom~\citep{Drautz2019,Batatia2022}. Section~\ref{sec:arch} specifies the cutoffs $l_q$ and maps $\mathcal W^{(q)}$, which define the fixed-width ladder in this work. Both operations use the same tensor products, but convolution sums contributions from neighbors, whereas the recurrence acts independently at each atom. This distinction motivates the two kernel designs in Section~\ref{sec:schedule}. Further details on the edge convolution and node-local products are provided in Appendix~\ref{app:background-details}.

\textbf{Irreducible Cartesian tensors.} For $l\ge0$, $\mathcal V_l\subset(\mathbb{R}^3)^{\otimes l}$ denotes the space of symmetric trace-free (STF) Cartesian tensors of rank $l$. These tensors are invariant under every permutation of their indices and, for $l\ge2$, have vanishing traces. Under $\bm R\in O(3)$, a tensor transforms as $\bm x_l\mapsto\bm R^{\otimes l}\bm x_l$, with $\bm R$ acting on each index. Inversion, i.e., $\bm R=-\bm I$, therefore contributes one factor of $-1$ per index, giving $\mathcal V_l$ natural parity of $(-1)^l$. A totally symmetric rank-$l$ tensor has $(l+1)(l+2)/2$ independent components, tracelessness imposes $l(l-1)/2$ conditions, and the difference $2l+1$ is the dimension of $\mathcal V_l$. Real spherical tensors of degree $l$ and rank-$l$ STF Cartesian tensors represent the same $O(3)$ irrep and are related by a fixed invertible equivariant map (Appendix~\ref{app:background-details}).

For $\lvert l_1-l_2\rvert\le l_3\le l_1+l_2$, the decomposition of $\mathcal V_{l_1}\otimes\mathcal V_{l_2}$ contains a single irreducible component labelled by $l_3$, with parity $(-1)^{l_1+l_2}$, which is natural exactly when $J=l_1+l_2+l_3$ is even. We call such paths $(l_1,l_2)\to l_3$ admissible. For an admissible path, the equivariant bilinear maps $\mathcal V_{l_1}\times\mathcal V_{l_2}\to\mathcal V_{l_3}$ form a one-dimensional space. Up to normalization, the spherical and Cartesian products therefore express the same coupling in different coordinates~\citep{Coope1970iii,Stone1976}. The representation thus affects the arithmetic and memory accesses needed to evaluate the coupling, but not its expressive power. Following prior work~\citep{Zaverkin2024, Lehman1989}, we write the Cartesian product along an admissible path as $(\bm x_{l_1}\hat{\otimes}_{\mathrm{Cart}}\bm y_{l_2})_{l_3}$. It contracts $n=(l_1+l_2-l_3)/2$ index pairs between the inputs, then symmetrizes the remaining $l_3$ indices and removes traces to obtain a rank-$l_3$ STF tensor (Appendix~\ref{app:background-details}). The next section expresses the natural-parity couplings in the $2l+1$ independent components.

\section{Generating Cartesian tensor products}
\label{sec:lowering}

We automatically generate GPU kernels for the Cartesian tensor products described in Section~\ref{sec:background} and their derivatives. For a chosen maximum tensor rank, we derive explicit symbolic expressions in independent components and translate them into Triton code~\citep{Tillet2019} for parallel execution on a GPU.

\textbf{Symbolic construction.} For each rank-$l$ tensor, we choose the $2l+1$ independent components containing at most one $z$ index, following Ref.~\citenum{Applequist1989} (Figure~\ref{fig:component-cost}(a)). Because the tensor is symmetric, each component is determined by the number of $x$, $y$, and $z$ indices. We denote these counts by $p_x$, $p_y$, and $p_z$, with $p_x+p_y+p_z=l$, and write the component as $(\bm x_l)_{p_xp_yp_z}$. The remaining components follow from the trace relation $(\bm x_l)_{p_xp_yp_z}=-(\bm x_l)_{p_x+2,p_y,p_z-2}-(\bm x_l)_{p_x,p_y+2,p_z-2}$ for $p_z\ge2$. Repeated application of this trace relation expresses every component of the rank-$l$ tensor through its $2l+1$ independent components. Our generator reconstructs all $3^l$ tensor components only symbolically, while model evaluation reads and writes the independent components.

For each admissible path $(l_1,l_2)\to l_3$, the generator reconstructs both inputs symbolically from their independent components and evaluates the Cartesian coupling described in Section~\ref{sec:background}. It computes their outer product, contracts $n=(l_1+l_2-l_3)/2$ index pairs between the inputs, and symmetrizes the remaining $l_3$ indices before removing traces. A path-dependent factor matches the normalization of the Cartesian product, accounting for the difference between averaging over index permutations and summing distinct index placements (Appendix~\ref{app:lowering-details}). For every admissible path up to the chosen maximum tensor rank, the generator selects the $2l_3+1$ independent components of the resulting tensor for code generation. Since these couplings are symmetric under exchange of the inputs, paths related by exchanging input ranks share one expression with exchanged operands.

Each independent output component is a bilinear polynomial in the independent input components with rational coefficients. For example, the $(1,1)\to2$ path gives $\bigl[(\bm x_1\hat{\otimes}_{\mathrm{Cart}}\bm y_1)_2\bigr]_{xx}=x_x y_x-(x_y y_y+x_z y_z)/2$. The generated code embeds the coefficients directly.

\textbf{Symbolic derivatives.} Training on forces requires differentiating through the VJP used to compute them. The polynomial expressions obtained above make both derivative operations accessible symbolically. For one coupling path, let $\bm f(\bm x,\bm y)$ denote the independent components of the unweighted output. We use SymPy to expand these expressions and differentiate them with respect to both inputs, obtaining explicit input VJPs. The tensor products used in the model also apply a scalar weight to each path and feature channel, so we require derivatives of the weighted contribution $\bm c(w,\bm x,\bm y)=w\bm f(\bm x,\bm y)$. Its input VJPs are scaled by $w$, while its weight VJP is the inner product of the incoming gradient with $\bm f(\bm x,\bm y)$. Because $\bm f$ is bilinear, differentiating this weighted VJP reuses the forward and input-VJP expressions, together with the terms arising from the weight dependence. Appendix~\ref{app:lowering-details} gives the resulting VJP and VJP-of-VJP formulas.

We embed these expressions in the Triton kernels described in Section~\ref{sec:schedule}. The kernels apply the scalar weights, optionally sum paths with the same output rank, and handle the input and output arrays for node-local products and convolutions over neighbors. All symbolic expansion and differentiation are completed before compilation. To verify the generated derivatives, we also generate a PyTorch forward implementation whose derivatives are computed by automatic differentiation.

\textbf{Cost of the generated expressions.} Expressing tensor products in independent components substantially reduces the arithmetic required compared with evaluating all $3^l$ Cartesian components. Counting additions and multiplications before compiler transformations, the operation count summed over all distinct couplings up to rank $3$ decreases from $1{,}299$ to $615$, close to the $609$ operations required by sparse Clebsch--Gordan coupling. Through ranks $4$ and $5$, the independent-component expressions require $1.3$--$1.5$ times as many operations as their spherical counterparts (Figure~\ref{fig:component-cost}(c)). The resulting Cartesian expressions therefore require a comparable number of arithmetic operations to sparse spherical coupling at the tested ranks. However, evaluation time also depends on memory access and kernel design, which we address in the next section.

\section{Executing Cartesian tensor products}
\label{sec:schedule}

The generated Triton kernels evaluate the symbolic expressions of Section~\ref{sec:lowering} for node-local tensor products and edge convolutions. For node-local products, each independently scheduled kernel instance, called a program, processes a group of atoms. The same implementation supports edge convolutions by gathering inputs along edges and atomically adding their contributions to receivers. Our default receiver-sorted kernel instead assigns each program a receiver, a tile of feature channels, and a group of output ranks. The program accumulates contributions from all incoming edges before storing the result, avoiding atomic additions in the forward pass. Here we focus on this kernel and adapt it for derivative evaluation, using atomic additions when multiple programs contribute to the same gradient. Appendix~\ref{app:schedule-details} outlines the implementation of the node-local kernels.

\textbf{Data layout.} Each row of an equivariant array contains the features of a node or an edge, with a contiguous block for each tensor rank. Within the block for tensor rank $l$, entries are ordered by the independent Cartesian component and then by the coupling path, with the feature channels for each pair stored contiguously. The blocks follow increasing tensor rank, so retaining only lower ranks requires no rearrangement or copying of the features. Figure~\ref{fig:layout-ownership} illustrates this layout, and Appendix~\ref{app:schedule-details} specifies how its entries are addressed in memory.

\textbf{Receiver-sorted forward.} For an edge $e$ from sender $v_e$ to receiver $u_e$, the kernel combines the projected sender features $\bm x_{v_e k}$ with the direction tensor $\bm y_e$, shared across channels. With a scalar weight $w_{e\eta k}$ for each edge, coupling path $\eta$, and channel $k$, the output is
\begin{equation}
  \bm z_{u\eta k}
  =\sum_{e:u_e=u} w_{e\eta k}\,
    \bm f_\eta(\bm x_{v_e k},\bm y_e),
  \label{eq:receiver-convolution}
\end{equation}
where $\bm f_\eta$ is the unweighted Cartesian product for the input and output ranks selected by $\eta$. This is the convolution of Section~\ref{sec:background} with separate outputs for each path. The neighbor normalization and channel mixing with $\bm W^{(t)}$ remain outside the kernel.

\begin{figure}[tbp]
\centering
\begin{minipage}[t]{0.48\textwidth}
\vspace{0pt}
\centering
\includegraphics[width=0.72\textwidth, trim=0 0pt 0 3pt, clip]{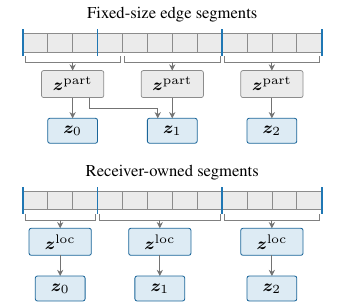}
\caption{\textbf{Accumulation over receiver-sorted edges.} Blue lines separate edges of different receivers. Fixed-size segments can split a receiver's sum across programs, while assigning all its edges to one program allows the sum to be computed locally and stored once.}
\label{fig:receiver-ownership}
\end{minipage}\hfill
\begin{minipage}[t]{0.50\textwidth}
\vspace{0pt}
\setlength{\intextsep}{0pt}
\begin{algorithm}[H]
\caption{Receiver-sorted forward kernel}
\label{alg:csr-fwd}
\fontsize{8.75}{10.5}\selectfont
\algrenewcommand\algorithmicindent{0.8em}
\begin{algorithmic}[1]
\Require Receiver-sorted arrays \texttt{rowptr} and \texttt{sender}
\Require Tensor inputs $\bm x,\bm y$ and scalar weights $w$
\Require Rank groups $\mathcal G$, paths $\mathcal P_{\mathcal G}$, and channel tiles $\mathcal K$
\For{each rank group $\mathcal G$}
  \ForAll{receivers $u$ and channel tiles $\mathcal K$}
    \State Load $e_0,e_1\gets\texttt{rowptr}[u],\texttt{rowptr}[u+1]$
    \State Initialize $\bm z^{\mathrm{loc}}_{\eta k}\gets\bm0$ for $\eta\in\mathcal P_{\mathcal G}$, $k\in\mathcal K$
    \For{$e=e_0,\ldots,e_1-1$}
      \State Load $v\gets\texttt{sender}[e]$, $\bm x_{v\mathcal K}$, and $\bm y_e$
      \For{$\eta\in\mathcal P_{\mathcal G}$, $k\in\mathcal K$}
        \State Load $w_{e\eta k}$
        \State $\bm z^{\mathrm{loc}}_{\eta k}
          \mathrel{+}=w_{e\eta k}\bm f_\eta(\bm x_{vk},\bm y_e)$
      \EndFor
    \EndFor
    \State Store $\bm z_{u\eta k}\gets\bm z^{\mathrm{loc}}_{\eta k}$ for $\eta\in\mathcal P_{\mathcal G}$, $k\in\mathcal K$
  \EndFor
\EndFor
\end{algorithmic}
\end{algorithm}
\end{minipage}
\end{figure}

We store edges in receiver order, with a row-pointer array locating each receiver's incoming edges and a sender array storing each edge's source-atom index. Each program evaluates the generated products over all incoming edges of one receiver for a channel tile $\mathcal K$ and a group of output ranks $\mathcal G$. It accumulates the sums locally and stores each once, without atomic additions or an intermediate tensor of edge messages. Separate kernel launches compute different groups of output ranks, limiting the number of sums accumulated simultaneously. Algorithm~\ref{alg:csr-fwd} describes the forward convolution, with $\mathcal P_{\mathcal G}$ denoting all coupling paths that produce tensor ranks in $\mathcal G$. Figure~\ref{fig:receiver-ownership} compares assigning all incoming edges of a receiver to one program with the fixed-size edge segments used by OpenEquivariance's deterministic forward kernel~\citep{Bharadwaj2025}.

\textbf{Receiver-sorted derivatives.} The VJP processes the same receiver-sorted edges and loads the receiver's incoming gradient $\bm g_{u\eta k}$ once per program. For each edge, it evaluates the generated input VJPs and computes the weight gradient by taking the inner product of $\bm g_{u\eta k}$ with the unweighted output $\bm f_\eta(\bm x_{v_e k},\bm y_e)$. Several receivers may contribute to the same sender gradient, while the direction-tensor gradient receives contributions from multiple feature channels and coupling paths. We accumulate shared input gradients using atomic additions, but store each weight gradient directly because each weight belongs to a single edge, coupling path, and feature channel. This also allows the VJP to divide a receiver's incoming edges into chunks processed by several programs, increasing parallelism for receivers with many neighbors without a separate step to combine partial receiver sums.

The VJP-of-VJP processes the same receiver-sorted edges using the generated forward and input-VJP expressions. Each program accumulates contributions from all incoming edges of receiver $u$ to compute $\widehat{\bm g}_{u\eta k}$, the gradient with respect to the incoming gradient $\bm g_{u\eta k}$, and stores the sum once. As in the VJP, we accumulate shared input gradients $\widehat{\bm x}$ and $\widehat{\bm y}$ using atomic additions, but store each weight gradient $\widehat w$ directly. Appendix~\ref{app:schedule-details} describes both derivative algorithms, including VJP edge chunking and additional kernel launches at higher tensor ranks.

\textbf{Comparison with other kernels.} In our benchmarks, we use OpenEquivariance's atomic mode~\citep{Bharadwaj2025}, which accumulates contributions to each receiver atomically in the forward pass. Its backward pass loads the receiver gradient for each edge and, like FlashCart, accumulates sender-feature gradients atomically. FlashCart groups edges by receiver, accumulating contributions in each forward program before storing the result and reusing each loaded receiver gradient across all edges in a VJP program. In OpenEquivariance's deterministic mode, fixed-size forward edge segments can split a receiver's sum across warps, requiring temporary storage and a second kernel to combine partial sums (Figure~\ref{fig:receiver-ownership}). Its backward pass uses an additional permutation to order edges by sender, which must be updated when the neighbor list changes~\citep{Bharadwaj2025}. Sobek likewise orders edges by sender to accumulate sender-feature gradients deterministically~\citep{Chorosajev2026}. FlashCart retains receiver ordering throughout, but its atomic gradient accumulation does not guarantee bitwise reproducibility.

\section{Model architecture}
\label{sec:arch}

FlashCart consists of $T$ message-passing blocks, each containing an equivariant convolution followed by a node-local product ladder. Here we describe the ladder and its channel mixing, with details of the convolution and the remaining architecture provided in Appendix~\ref{app:architecture-details}.

\textbf{Factorized fixed-width correlations.} The recursive construction of Section~\ref{sec:background} leaves open which ranks are retained after each product and how the channels are mixed. We propose a fixed-width ladder using channel-diagonal Cartesian products, which couple only input features with the same channel index. A learned linear map then mixes paths and channels to produce $d$ channels per retained tensor rank. Each additional correlation order requires one pairwise product and one linear map, using the generated Cartesian kernels without a separate higher-order tensor-product operator.

We initialize the ladder from the atomic basis $\bm A_{ukl}$ as $\bm\phi^{(1)}_{ukl}=\sum_{k'} W^{(1)}_{l,kk'}\bm A_{uk'l}$, where the learned $d\times d$ map $W^{(1)}$ mixes channels independently at each tensor rank. For $q=2,\ldots,\nu$, each step applies a Cartesian product followed by learned channel mixing,
\begin{equation}
  \bm B^{(q)}_{u\eta_qkl}
  =\left(\bm\phi^{(q-1)}_{ukl_1}
      \hat{\otimes}_{\mathrm{Cart}}\bm A_{ukl_2}\right)_l,
  \quad
  \bm\phi^{(q)}_{ukl}
  =\sum\nolimits_{\eta_q,k'}W^{(q)}_{l,k(\eta_q k')}\bm B^{(q)}_{u\eta_qk'l},
  \qquad l\le l_q.
  \label{eq:ladder}
\end{equation}
Here $\eta_q=(l_1,l_2\to l)$ is an admissible coupling path, and the cutoff $l_q$ specifies the highest output rank retained at step $q$. For $q<\nu$, each step retains tensor ranks up to $l_q=l_{\max}^{\mathrm{prod}}$. The last step retains ranks up to $l_\nu=l_{\max}^{\mathrm{out}}$, which equals $l_{\max}^{\mathrm{prop}}$ in all blocks except the final one, where it is zero and only scalars are returned. Summing the contributions from all steps gives $\bm z_u=\sum_{q=1}^{\nu}\Pi_{\le l_{\max}^{\mathrm{out}}}\bm\phi^{(q)}_u$, where the projection retains only the tensor ranks required at the ladder output. Each product introduces another factor of $\bm A$, making $\bm\phi^{(q)}$ a polynomial of degree $q$ in the basis. Because normalization and nonlinear gating make $\bm A$ depend nonlinearly on the neighborhood, $q$ specifies the polynomial degree in $\bm A$, not a bound on the number of interacting neighbors.

In Equation~\ref{eq:ladder}, the product couples channel $k$ of $\bm\phi^{(q-1)}$ with channel $k$ of $\bm A$. The learned map $W^{(q)}$ then mixes the resulting paths and channels separately for each output tensor rank. With $P_q$ retained paths, this gives $P_qd^2$ learned parameters per step, compared with $P_qd^3$ for a dense map combining products between all pairs of input channels. Each step returns $d$ channels per retained tensor rank, allowing the ladder to build higher-order correlations without increasing its channel width. The ladder's learned mixing maps $W^{(q)}$ contain $d^2\sum_{q=1}^{\nu}P_q$ parameters in total. For the initialization, $P_1$ is the number of retained tensor ranks, each with its own $d\times d$ map. With the maximum tensor ranks held fixed, both the parameter count and the computational cost per atom grow quadratically with the channel width $d$ and linearly with the correlation degree $\nu$.

Our node-local correlation ladder combines compression between successive correlation orders, as in NICE and TRACE~\citep{Nigam2020,Ahlawat2026}, with channel-diagonal products followed by dense learned mixing, as used in Cormorant and Allegro~\citep{Anderson2019,Musaelian2023}. Our ladder reuses the same atomic basis at every step. Expanding this recurrence gives a tensor-train factorization of the polynomial coefficients for each retained sequence of coupling paths, with the $d$ channels passed between steps connecting its factors (Appendix~\ref{app:architecture-details}). We also evaluate path-reduced products that sum weighted paths at each step and mix channels only after the recurrence. The smallest model in Figure~\ref{fig:pareto} uses these products, which are also evaluated alongside the full ladder in the node-local kernel benchmark in Section~\ref{sec:results}.

\section{Results and discussion}
\label{sec:results}

We benchmark the generated kernels, examine how architectural choices affect the tradeoff between accuracy and computational cost, and compare FlashCart models with state-of-the-art potentials on the SPICE-MACE-OFF data set~\citep{Kovacs2025}. We also evaluate a FlashCart model trained on the MAD data set~\citep{Mazitov2025} across a wider range of elements and atomic environments, comparing its accuracy with PET-MAD~\citep{Mazitov2025b}, a state-of-the-art transformer-based potential. We measure runtime on an NVIDIA RTX PRO 6000 Blackwell GPU (96~GB) and report FlashCart prediction errors averaged over three independent runs. Appendix~\ref{app:protocol} provides further details on the datasets, training, and benchmarks.

\begin{figure}[t]
\centering
\includegraphics[width=0.95\textwidth, trim=0 10pt 0 6.5pt, clip]{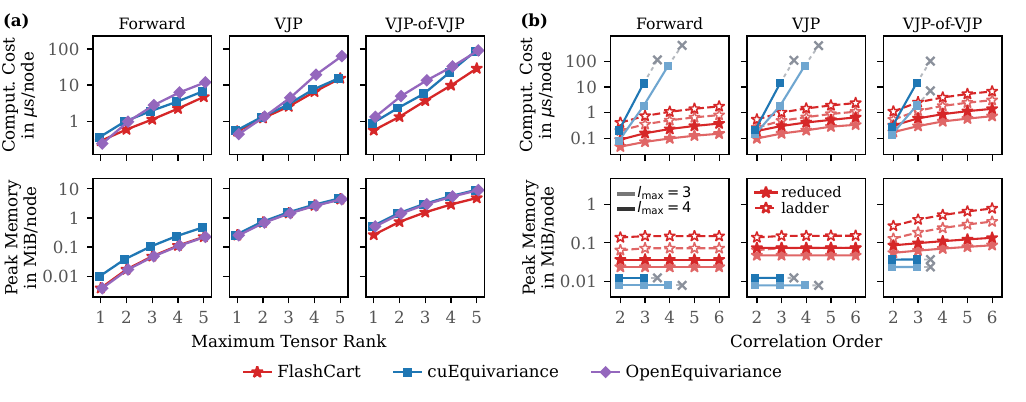}
\caption{\textbf{Runtime and memory of edge convolutions and node-local tensor products.} (a)~Results for edge convolutions with $1024$ nodes and $128$ edges per node as a function of the maximum tensor rank. (b)~Results for node-local tensor products with $4096$ nodes as a function of the correlation order. Solid FlashCart curves show path-reduced products that sum contributions within the kernel, with a scalar weight per path and feature channel. Dashed curves show the full ladder, including the linear maps that mix paths and channels between steps. cuEquivariance evaluates symmetric contractions with a learned weight for each contraction term and feature channel. Both implementations use the indicated maximum rank for their input and output tensors. Grey crosses denote cases where contraction construction was not completed, or a kernel failed to launch (Appendix~\ref{app:additional-results}). Dashed extensions indicate trends beyond the available measurements. All results are averaged over three independent runs and use $128$ feature channels and single precision.}
\label{fig:kernel-results}
\end{figure}

\textbf{Kernel microbenchmarks.} Figure~\ref{fig:kernel-results}(a) compares edge convolutions with $1024$ nodes, $128$ feature channels, and $128$ neighbors per node as the maximum tensor rank increases from $1$ to $5$. Over the full benchmark grid of $30$ configurations, which also includes $64$ and $256$ channels and $64$ neighbors per node, FlashCart is on average $1.6\times$, $1.1\times$, and $1.9\times$ faster than cuEquivariance for the forward, VJP, and VJP-of-VJP, respectively. Compared with OpenEquivariance, FlashCart is $1.9\times$, $2.1\times$, and $3.1\times$ faster on average, although OpenEquivariance is faster in some configurations at ranks $\{1, 2\}$. The largest average improvement occurs in the VJP-of-VJP used to differentiate the force loss. In this phase, FlashCart also uses about half the peak GPU memory of either baseline. These timings compare the complete kernel implementations, including memory access and the aggregation of edge contributions. The operation counts in Section~\ref{sec:lowering} show no arithmetic advantage over sparse spherical coupling and therefore do not explain the observed runtime advantages.

Figure~\ref{fig:kernel-results}(b) compares node-local products with $4096$ nodes, $128$ feature channels, and maximum tensor ranks $3$ and $4$. At fixed tensor ranks and width, each increase in correlation order adds one pairwise product to the FlashCart recurrence. The full ladder also applies a learned linear map after each product, whose runtime and memory are included in the measurements. At order $2$, path-reduced products have comparable runtime to cuEquivariance's symmetric contraction, while the full ladder is slower. At higher orders, both FlashCart variants become faster than cuEquivariance. At rank $3$ and order $4$, forward evaluation takes $0.40$~ms for path-reduced products and $2.07$~ms for the full ladder, compared with $280.61$~ms for cuEquivariance, making them approximately $695\times$ and $136\times$ faster, respectively. Furthermore, FlashCart reaches order $6$ at every tested rank, while some cuEquivariance contractions could not be constructed or their kernels failed to launch (Appendix~\ref{app:additional-results}).

We note that reported runtime differences also reflect how the correlations are constructed. cuEquivariance assigns learned weights to complete symmetric contractions, while FlashCart applies weighted products successively, retaining only tensor ranks up to the chosen maximum after each step. Within FlashCart, the full ladder retains separate path contributions for learned mixing, which requires more memory than summing them within the product kernel. Its additional peak GPU allocation reaches $3.1$~GiB for the VJP-of-VJP at rank $4$ and order $6$.

\begin{figure}[t]
\centering
\includegraphics[width=0.95\textwidth, trim=0 6pt 0 14pt, clip]{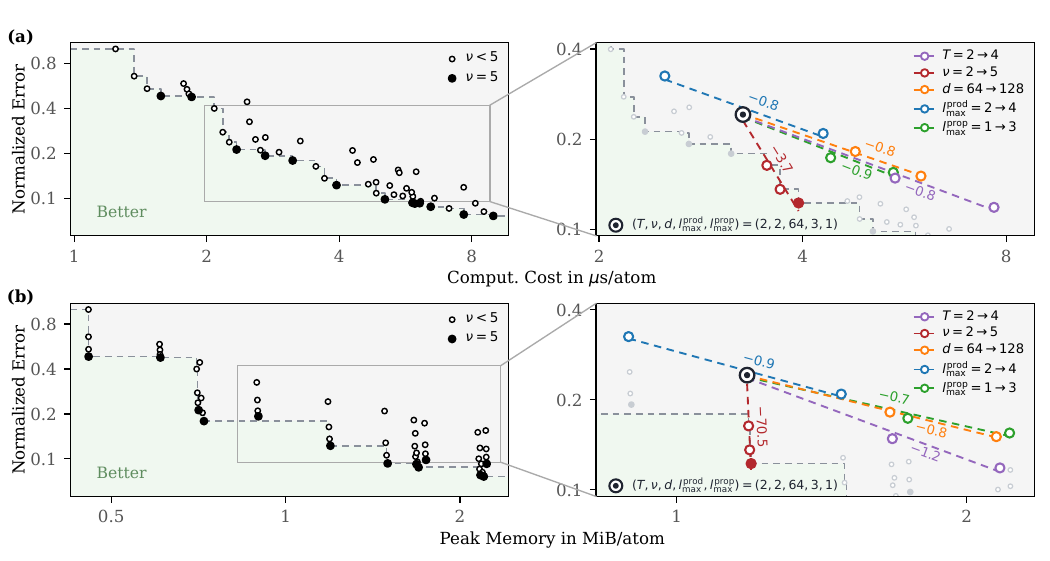}
\caption{\textbf{Accuracy--efficiency tradeoffs across architectural choices.} Normalized error on SPICE-MACE-OFF is shown against runtime (a) and peak GPU memory (b). Filled symbols denote $\nu=5$, while unfilled ones denote lower correlation orders. The insets show the effect of varying one architectural choice at a time, with the maximum edge and product ranks varied together. Larger negative fitted slopes indicate larger relative error reductions per relative increase in runtime or memory. We average errors over three independent runs and measure runtime and memory for energy and force evaluation on $4096$ diamond atoms using one trained model per configuration.}
\label{fig:efficiency}
\end{figure}

\textbf{Efficiency of higher-order correlations.} We next compare the accuracy and cost of increasing correlation order with those of increasing channel width, the number of message-passing blocks, or maximum tensor ranks (Figure~\ref{fig:efficiency}). We quantify accuracy using normalized error, which averages the energy and force MAEs after dividing each by a fixed reference value (Appendix~\ref{app:protocol}). On the logarithmic axes, the fitted slope describes the relative change in error for a relative change in runtime or memory. For example, a slope of $-3$ in the runtime panel corresponds to approximately $3\%$ lower error for $1\%$ more runtime along the fitted curve. Across the tested configurations, runtime slopes range from $-6.5$ to $-2.6$ for correlation order and from $-1.5$ to $-0.8$ for the other architectural choices. Increasing correlation order therefore gives the largest error reduction relative to added runtime. Across configurations with the remaining architecture fixed, raising $\nu$ from $2$ to $5$ reduces normalized error by $19$ to $60\%$, while runtime increases by $4$ to $37\%$ and peak GPU memory by at most $1.7\%$. The steep memory slopes reflect substantial error reductions with little change in memory use. Consistent with these results, most models on the observed accuracy--efficiency frontier have $\nu=4$ and $5$, for both runtime and memory.

Figure~\ref{fig:ablation} separates the results of Figure~\ref{fig:efficiency} by tensor rank, channel width, and number of message-passing blocks, showing how energy error, force error, and runtime vary with correlation order. When both the direction tensors and the intermediate tensors in the correlation ladder are limited to rank $1$, raising $\nu$ from $2$ to $5$ lowers energy error by $20\%$ and increases runtime by $4\%$. At maximum ranks of $4$, the corresponding changes are $58\%$ and $37\%$. Higher tensor ranks thus allow larger relative error reductions from increasing order, at greater computational cost. The relative reductions in energy and force errors from increasing correlation order generally become smaller with more message-passing blocks or a higher maximum propagated rank, but vary little with channel width. Since increasing order also adds parameters, we additionally compare models with approximately $0.93$M parameters, reducing channel width as $\nu$ increases. At maximum edge and product rank $3$, an order-$5$ model with $64$ channels reaches a force MAE of $8.29$~meV/\AA\ at $3.94$~$\mu$s/atom, compared with $11.53$~meV/\AA\ at $5.49$~$\mu$s/atom for order $2$ with $108$ channels. For maximum edge and product ranks $1$ to $4$, increasing order while reducing channel width lowers force error by $9$ to $29\%$ and runtime by $10$ to $34\%$ at approximately the same parameter count.

\textbf{Model accuracy and computational cost.} Figure~\ref{fig:pareto} compares five FlashCart models with state-of-the-art MLIPs using a normalized error that combines energy and force MAEs averaged equally over the seven SPICE test subsets. Table~\ref{tab:pareto-data} reports the separate energy and force errors for each subset. Among models benchmarked on the same GPU, the FlashCart models lie on the observed accuracy--efficiency frontier for both runtime and memory. The largest FlashCart model has $5.6$M parameters and reaches energy and force MAEs of $0.072\pm0.002$~meV/atom and $1.99\pm0.04$~meV/\AA. Evaluating energies and forces for $1000$ diamond atoms takes $21.5$~$\mu$s/atom and $5.2$~MiB/atom of peak GPU memory. This model improves on PET-L's errors of $0.106$~meV/atom and $2.52$~meV/\AA, despite having far fewer parameters and running approximately ten times faster. It also approaches DPA4-Plus' force MAE of $1.80$~meV/\AA\ while improving on its energy MAE of $0.091$~meV/atom, with $7.7\times$ faster evaluation and $3.1\times$ less memory. The $5.6$M model thus approaches the lowest reported force errors while achieving lower combined error with substantially less runtime and memory.

The timings use each model's own cutoff radius. In diamond, FlashCart processes $158$ edges per atom compared with PET-L's $70$, giving it an approximately $23\times$ runtime advantage per edge. This normalization accounts for the different numbers of neighbors, although the comparison still includes architectural differences. Runtime per atom also depends on system size, approaching a constant at smaller atom counts for the larger FlashCart models than for the smallest one (Figure~\ref{fig:pareto}(b)).

To measure the contribution of the generated kernels, Figure~\ref{fig:backend} compares them with the corresponding PyTorch implementations of tensor products and edge convolutions, using the same $5.6$M model checkpoint. With both implementations compiled, the generated kernels reduce inference time from $175.5$ to $21.5$~$\mu$s/atom and runtime in the training benchmark from $442.6$ to $82.2$~$\mu$s/atom, making them $8.2\times$ and $5.4\times$ faster, respectively. Compilation also accelerates the generated implementation itself, while increasing its peak GPU memory during training by $24\%$ (Appendix~\ref{app:additional-results}).

\begin{table}[t]
\centering
\caption{\textbf{Energy and force errors on the PET-MAD benchmark.} Each energy/force pair gives MAEs in meV/atom and meV/\AA, respectively. FlashCart and PET-MAD are trained on MAD for $400$ and $1500$ epochs with $5.0$ and $4.5$~\AA\ cutoffs, respectively. Both use reference energies and forces computed at MAD's electronic-structure level. The other baselines use reference calculations consistent with their training data. FlashCart results show the mean and standard deviation across three random seeds with a fixed data split. The lowest energy and force MAEs are shown in bold.}
\label{tab:mad}
\setlength{\tabcolsep}{3pt}
\begin{adjustbox}{max width=\textwidth}
\begin{tabular}{
    l@{\hspace{4pt}}
    w{r}{4.35em}@{\,/\,}w{l}{4.35em}@{\hspace{4pt}}
    w{r}{2.85em}@{\,/\,}w{l}{2.85em}@{\hspace{4pt}}
    w{r}{3.9em}@{\,/\,}w{l}{3.9em}@{\hspace{4pt}}
    w{r}{3.7em}@{\,/\,}w{l}{3.7em}@{\hspace{4pt}}
    w{r}{2.3em}@{\,/\,}w{l}{2.3em}@{\hspace{4pt}}
    w{r}{3.55em}@{\,/\,}w{l}{3.55em}@{\hspace{4pt}}
}
\toprule
Dataset & \multicolumn{2}{c@{\hspace{4pt}}}{FlashCart} & \multicolumn{2}{c@{\hspace{4pt}}}{PET-MAD~\citep{Mazitov2025b}} & \multicolumn{2}{c@{\hspace{4pt}}}{MACE-MP-0-L~\citep{Batatia2025b}} & \multicolumn{2}{c@{\hspace{4pt}}}{MatterSim-5M~\citep{Yang2024}} & \multicolumn{2}{c@{\hspace{4pt}}}{Orb-v2~\citep{Neumann2024}} & \multicolumn{2}{c@{\hspace{4pt}}}{SevenNet-l3i5~\citep{Park2024}} \\
\cmidrule(r){1-1}\cmidrule(lr){2-5}\cmidrule(l){6-13}
MAD & \textbf{10.2}$\bm{\pm}$\textbf{0.4} & \textbf{52.7}$\bm{\pm}$\textbf{1.8} & 17.6 & 65.1 & 81.6 & 181.5 & 47.3 & 133.7 & 52.9 & 96.2 & 82.1 & 173.5 \\
MPtrj & 21.1$\pm$0.3 & 67.9$\pm$0.8 & 22.3 & 77.9 & 15.1 & 50.8 & 21.3 & 61.4 & \textbf{5.6} & \textbf{21.9} & 9.8 & 25.5 \\
MatBench & \textbf{29.6}$\bm{\pm}$\textbf{0.6} & -- & 31.3 & -- & 58.5 & -- & 38.2 & -- & 37.9 & -- & 47.5 & -- \\
Alexandria & 45.3$\pm$0.8 & 60.9$\pm$1.9 & 49.0 & 66.8 & 65.4 & 79.5 & 21.2 & 39.9 & \textbf{13.2} & \textbf{10.5} & 47.6 & 70.3 \\
OC2020 & \textbf{12.8}$\bm{\pm}$\textbf{0.3} & 103.9$\pm$1.1 & 18.3 & 114.5 & 82.4 & 169.6 & 31.5 & 119.2 & 19.8 & \textbf{99.3} & 45.7 & 162.7 \\
SPICE & \textbf{2.9}$\bm{\pm}$\textbf{0.3} & \textbf{43.3}$\bm{\pm}$\textbf{1.9} & 3.7 & 59.5 & 10.6 & 166.8 & 21.3 & 145.6 & 59.0 & 140.8 & 11.3 & 139.1 \\
MD22 & 3.4$\pm$0.4 & \textbf{54.2}$\bm{\pm}$\textbf{3.9} & \textbf{1.9} & 65.6 & 9.4 & 182.9 & 28.6 & 160.4 & 174.3 & 220.7 & 11.1 & 146.2 \\
\bottomrule
\end{tabular}
\end{adjustbox}
\end{table}

\textbf{Accuracy across materials and molecules.} We evaluate a $2.3$M-parameter FlashCart model trained on MAD~\cite{Mazitov2025}, which contains fewer than $100{,}000$ structures spanning $85$ elements. Table~\ref{tab:mad} compares it with the $2.8$M-parameter PET-MAD potential and other MLIPs on inorganic materials (MPtrj~\citep{Deng2023}, Matbench Discovery~\citep{Riebesell2025}, and Alexandria~\citep{Schmidt2023}), molecules adsorbed on catalyst surfaces (OC2020~\citep{Chanussot2021}), molecular configurations (SPICE~\citep{Eastman2023} and MD22~\citep{Chmiela2023}), and $360$ MAD test structures. FlashCart has lower force MAEs than PET-MAD on all six benchmarks reporting force errors and lower energy MAEs on six of seven benchmarks. Both MAD-trained models have lower energy and force errors on MAD and the molecular benchmarks than the other baselines. On the full MAD test set, FlashCart reaches energy and force MAEs of $11.4\pm0.1$~meV/atom and $63.9\pm0.5$~meV/\AA, compared with PET-MAD's $14.7$~meV/atom and $72.9$~meV/\AA\ (Table~\ref{tab:mad-test}). For energy and force evaluation on $1000$ diamond atoms, FlashCart and PET-MAD take $8.4$ and $45.1$~$\mu$s/atom and use $2.0$ and $3.3$~MiB/atom of peak GPU memory, respectively. These results establish FlashCart as a competitive architecture for general-purpose potentials, achieving accuracy comparable to state-of-the-art models across diverse materials and molecular environments.

\section{Conclusions}

FlashCart combines symbolic Cartesian tensor products with generated GPU kernels and a correlation ladder whose cost grows linearly with correlation order at fixed width and maximum tensor rank. These generated kernels often outperform optimized spherical counterparts in our benchmarks and reduce the time required for energy and force evaluation and differentiation of the training loss compared with a compiled PyTorch implementation of the same model. On SPICE-MACE-OFF, FlashCart models lie on the accuracy--efficiency frontier among models benchmarked on the same GPU. We also demonstrate that increasing correlation order improves the accuracy--runtime tradeoff more than increasing width, depth, or tensor rank in the tested configurations. Trading width for higher order also lowers force error and runtime at approximately fixed parameter counts. Finally, MAD results show competitive accuracy across diverse materials and molecular environments.

\textbf{Limitations.} FlashCart uses natural-parity tensor products, which can be extended to opposite parity through Levi--Civita contractions and parity labels. Fixed channel widths and truncated tensor ranks keep the ladder's intermediate features compact but limit the correlations it can represent. Retaining higher ranks also increases the size of the generated kernel code and its compilation cost.

\subsubsection*{Ethics statement}

This work develops computational methods for atomistic simulations using existing datasets. It does not involve human participants or personal data.

\subsubsection*{Reproducibility statement}

Appendix~\ref{app:protocol} describes the datasets and explains how we train and evaluate the models and benchmark the kernels. Source code is available on GitHub at \url{https://github.com/inm-leibniz/flashcart}. All trained FlashCart models are available on Zenodo at \url{https://doi.org/10.5281/zenodo.23159584}.

\subsubsection*{Acknowledgments}

MN acknowledges support from the Deutsche Forschungsgemeinschaft (DFG, German Research Foundation) through project number 569019417 and under Germany’s Excellence Strategy – EXC 2075 – 390740016, as well as from the Stuttgart Center for Simulation Science (SimTech). The authors acknowledge the use of OpenAI Codex and Anthropic Claude Code for assistance with code generation and manuscript preparation. Responsibility for the final manuscript and code remains with the authors.

\bibliography{references}
\bibliographystyle{bibstyle}

\clearpage
\appendix

\numberwithin{figure}{section}
\numberwithin{table}{section}
\numberwithin{equation}{section}
\renewcommand{\thefigure}{\thesection\arabic{figure}}
\renewcommand{\thetable}{\thesection\arabic{table}}
\renewcommand{\theequation}{\thesection\arabic{equation}}
\renewcommand{\theHfigure}{appendix.\thesection.\arabic{figure}}
\renewcommand{\theHtable}{appendix.\thesection.\arabic{table}}
\renewcommand{\theHequation}{appendix.\thesection.\arabic{equation}}
\section*{\texorpdfstring{\Large Appendices}{Appendices}}

\section{Related work}
\label{app:related-details}

\textbf{Product bases and recursive evaluation.} The atomic cluster expansion provides a body-ordered polynomial basis. It uses the density trick to evaluate many-body features through products of an atomic basis constructed by one-particle neighbor sums~\citep{Drautz2019}. MACE constructs equivariant many-body features by contracting products of the atomic basis with precomputed generalized Clebsch--Gordan tensors and learned weights~\citep{Batatia2022}. Recursive evaluation builds higher-order products from lower-order intermediates, reusing them to reduce the cost of constructing a product basis~\citep{Lysogorskiy2021}. ICTP uses a recursive construction for Cartesian correlations, combining the atomic basis through successive pairwise tensor products~\citep{Zaverkin2024}. The number of intermediate features can further be limited by compressing the output of each recursive step. NICE uses principal component analysis to reduce the number of equivariant features between successive correlation orders~\citep{Nigam2020}, while TRACE incorporates learned mixing within each tensor product, with weights that combine pairs of input channels into the retained output channels~\citep{Ahlawat2026}.

\textbf{Channel mixing and coefficient factorization.} Learned mixing reduces tensor-product outputs to the desired number of channels for each tensor rank, as in Clebsch--Gordan networks~\citep{Kondor2018}. Cormorant combines this mixing with channel-diagonal tensor products, which reduce the computational cost by coupling only matching input channels~\citep{Anderson2019}. Allegro uses this combination in its edgewise tensor-product layers, with dense mixing over both coupling paths and channels~\citep{Musaelian2023}. Channel-diagonal products also arise from factorizations of polynomial coefficients. Tensor-reduced ACE expresses these coefficients as sums of rank-one terms over chemical and radial indices, leading to products that share a channel index, as in MACE~\citep{Darby2023, Batatia2022}. GRACE shows how this shared-index factorization can be written as a tensor train whose intermediate cores preserve that index~\citep{Bochkarev2024, Oseledets2011}. Tensor trains also provide a direct parameterization of polynomial coefficients in equivariant tensor network potentials~\citep{Hodapp2024}.

\textbf{Irreducible Cartesian tensors.} The description of irreducible Cartesian tensors in Section~\ref{sec:background} and Appendix~\ref{app:background-details} follows classical work on irreducible tensors~\citep{Coope1965, Coope1970ii, Coope1970iii, Stone1976, Applequist1989, Lehman1989,Snider2018}. ICTP provides the framework for equivariant message passing with higher-rank irreducible Cartesian tensors that we follow, including its tensor-product conventions~\citep{Zaverkin2024}. Other Cartesian potentials include TensorNet with rank-two tensors and HotPP with reducible higher-rank tensors~\citep{Simeon2023, Wang2024}. Cartesian-3j implements Cartesian couplings through transformed Wigner-$3j$ coefficients~\citep{Xu2026}, while CarNet provides precomputed projectors for natural tensors, their products, and general tensorial observables~\citep{Chen2026}.

\textbf{Accelerated tensor-product kernels.} For spherical tensor representations, cuEquivariance, OpenEquivariance, and Sobek accelerate tensor products and convolutions with specialized GPU kernels~\citep{Geiger2024, Bharadwaj2025, Chorosajev2026}. Optimizations across these libraries include exploiting sparse coupling coefficients and fusing tensor products with neighbor aggregation to reduce intermediate storage and memory traffic. cuEquivariance describes these operations through segmented tensor products, which specify blocks of tensor components and the contractions between them~\citep{Geiger2024}. OpenEquivariance generates specialized GPU kernels for the requested Clebsch--Gordan couplings, incorporating their nonzero coefficients directly into the generated code~\citep{Bharadwaj2025}. The Triton implementation for Allegro evaluates tensor products using only the nonzero Wigner-$3j$ coefficients and their corresponding component indices~\citep{Tan2026}. Sobek supports dense and channel-wise mixing and reorders the convolution to perform receiver accumulation before channel matrix products, avoiding intermediate edge-weight tensors~\citep{Chorosajev2026}.

\section{Background}
\label{app:background-details}

\textbf{Equivariant convolution and many-body correlations.} Equivariant architectures differ in the body order represented within a local neighborhood, the angular ranks retained at different layers, the number of feature channels, and their depth~\citep{Batatia2025, Joshi2023}. A radial envelope makes every edge contribution vanish smoothly at the cutoff, so the energy remains twice differentiable in the positions. Suppressing channel indices, the convolution used in this work reads as
\begin{equation*}
  \bm A_{ul}^{(t)}
  =\sum_{v\in\mathcal N(u)}
   \sum_{\substack{(l_1,l_2)\to l\\[-1pt]
                     l_1\le l_{\max}^{\mathrm{prop}},\;
                     l_2\le l_{\max}^{\mathrm{edge}}}}
   \bm W_{l_1l_2l}^{(t)}(r_{uv})
   \bigl(\bm h_{vl_1}^{(t)}\hat{\otimes}\bm T_{l_2}(\hat{\bm r}_{uv})\bigr)_l,
   \qquad l\le l_{\max}^{\mathrm{prod}}.
\end{equation*}
Here $(l_1,l_2)\to l$ is an admissible coupling path, and the distance-dependent maps $\bm W_{l_1l_2l}^{(t)}$ combine radial weighting with channel mixing for each tensor rank. The product $\hat{\otimes}$ denotes equivariant coupling, with $(\cdot)_l$ selecting the output rank. It is evaluated using Clebsch--Gordan coefficients in the spherical basis or the Cartesian contractions defined below~\citep{Zaverkin2024}.

Writing $\bm A_u=\bigoplus_l\bm A_{ul}$, the node-local products are evaluated recursively as
\begin{equation*}
  \bm\phi_u^{(0)}=1,\qquad
  \bm\phi_u^{(q)}
  =\mathcal W^{(q)}\Pi_{\le l_q}
    \bigl(\bm\phi_u^{(q-1)}\hat{\otimes}\bm A_u\bigr),
  \qquad q=1,\ldots,\nu.
\end{equation*}
The projection $\Pi_{\le l_q}$ discards ranks above $l_q$, the maps $\mathcal W^{(q)}$ are equivariant and linear, and every step contributes to the layer output. Equivariance fixes neither the retained ranks nor the form of these maps. Since $\bm\phi_u^{(q)}$ is degree $q$ in $\bm A_u$, expanding the neighbor sums gives local terms with at most $q$ neighbors. Counting the central atom gives a maximum body order of $q+1$, with lower-order terms when neighbor indices repeat~\citep{Drautz2019, Batatia2022}. Stacking $T$ message-passing layers extends the effective interaction range to $Tr_{\max}$.

\textbf{Irreducible Cartesian tensors.} The real spherical tensor space $\mathcal H_l\cong\mathbb R^{2l+1}$ and the Cartesian STF space $\mathcal V_l$ represent the same $O(3)$ irrep. An invertible equivariant map $B_l:\mathcal H_l\to\mathcal V_l$ therefore exists and is unique up to a nonzero scalar. For each admissible path, the equivariant bilinear coupling is also unique up to normalization, so the two products satisfy~\citep{Coope1970iii,Stone1976}
\begin{equation*}
  (\bm x_{l_1}\hat{\otimes}_{\mathrm{Cart}}\bm y_{l_2})_{l_3}
  =\kappa_{l_1l_2l_3}\,
   B_{l_3}\bigl[\bigl(B_{l_1}^{-1}\bm x_{l_1}\hat{\otimes}_{\mathrm{CG}}
   B_{l_2}^{-1}\bm y_{l_2}\bigr)_{l_3}\bigr],
\end{equation*}
where $\kappa_{l_1l_2l_3}$ accounts for the product normalizations and the scaling of $B_l$.

The direction tensors used in the convolution are normalized STF projections of repeated outer products of the unit direction. Following Refs.~\citenum{Zaverkin2024, Lehman1989}, we define
\begin{equation*}
  \bm T_l(\hat{\bm r}_{uv}) =
  \frac{(2l-1)!!}{l!}\sum_{m=0}^{\lfloor l/2\rfloor}
  (-1)^m\frac{(2l-2m-1)!!}{(2l-1)!!}
  \{\hat{\bm r}_{uv}^{\,\otimes(l-2m)}\otimes\bm I^{\otimes m}\}.
\end{equation*}
Here $\bm I$ is the identity tensor, $(-1)!!=1$, and the braces sum over distinct placements of the free indices, leading to $l!/[(l-2m)!\,2^m\,m!]$ terms. For example, $\{\hat{\bm r}\otimes\bm I\}_{ijk}=\hat r_i\delta_{jk}+\hat r_j\delta_{ik}+\hat r_k\delta_{ij}$. The prefactor normalizes the full contraction of $\bm T_l$ with $\hat{\bm r}_{uv}$ to unity, and the $\bm I$ terms remove traces while preserving rank $l$. We collect all edge ranks as $\bm T(\hat{\bm r}_{uv})=\bigoplus_{l=0}^{l_{\max}^{\mathrm{edge}}}\bm T_l(\hat{\bm r}_{uv})$.

The same symmetrization and trace removal define products between two STF tensors. For an admissible path $(l_1,l_2)\to l_3$, $J=l_1+l_2+l_3$ is even and contracting $n=(l_1+l_2-l_3)/2$ index pairs reduces the total rank to $l_3$. With $m_{\max}=\min(l_1,l_2)-n$, the bounds $n\ge0$ and $m_{\max}\ge0$ are equivalent to the triangle inequalities. For $\bm x_{l_1}\in\mathcal V_{l_1}$ and $\bm y_{l_2}\in\mathcal V_{l_2}$, the coupling reads~\citep{Zaverkin2024, Lehman1989}
\begin{equation}
  (\bm x_{l_1}\hat{\otimes}_{\mathrm{Cart}}\bm y_{l_2})_{l_3}
  = C_{l_1l_2l_3}\sum_{m=0}^{m_{\max}}
  (-2)^m\frac{(2l_3-2m-1)!!}{(2l_3-1)!!}
  \{(\bm x_{l_1}\cdot(n{+}m)\cdot\bm y_{l_2})
  \otimes\bm I^{\otimes m}\}.
  \label{eq:tp}
\end{equation}
Here $\bm x_{l_1}\cdot q\cdot\bm y_{l_2}$ contracts $q$ index pairs by summing over each shared Cartesian index. The $m$-th term contracts $n+m$ pairs and introduces $m$ identity tensors to restore rank $l_3$. The braces sum over distinct placements of the free indices among the remaining input indices and the identity factors. The constant $C_{l_1l_2l_3}$ fixes the path normalization and is defined elsewhere~\citep{Zaverkin2024}.

Paths with odd $J$ require the Levi--Civita symbol and produce outputs with parity opposite to $(-1)^{l_3}$. We restrict this work to natural-parity outputs, but the same symbolic construction extends to odd-$J$ couplings by including Levi--Civita contractions and explicitly tracking parity.

\section{Generating Cartesian tensor products}
\label{app:lowering-details}

\begin{figure}[t]
  \centering
  \begin{minipage}[t]{0.98\textwidth}
    \vspace{0pt}
    \centering
    \includegraphics[width=\linewidth]{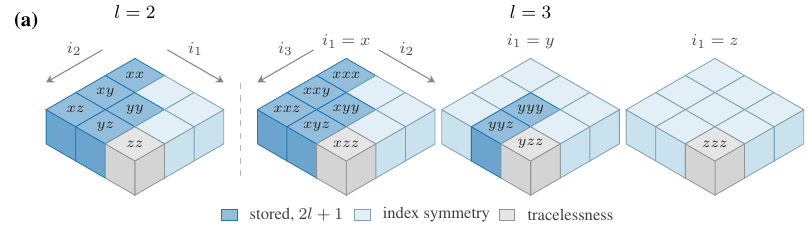}
  \end{minipage}
  \par\vspace{1.5mm}
  \begin{minipage}[t]{0.98\textwidth}
    \centering
    \begin{minipage}[t]{0.24\linewidth}
      \vspace{0pt}
      \centering
      \includegraphics[width=\linewidth]{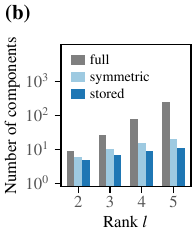}
    \end{minipage}\hfill
    \begin{minipage}[t]{0.51\linewidth}
      \vspace{0pt}
      \centering
      \includegraphics[width=\linewidth]{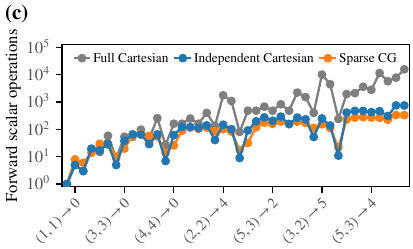}
    \end{minipage}\hfill
    \begin{minipage}[t]{0.24\linewidth}
      \vspace{0pt}
      \centering
      \includegraphics[width=\linewidth]{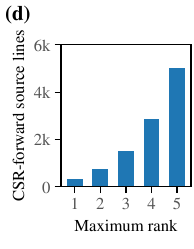}
    \end{minipage}
  \end{minipage}
  \caption{\textbf{Independent Cartesian components and the cost of generated tensor products.} (a)~Stored components for ranks $2$ and $3$ are blue, with light blue entries obtained by index symmetry and grey entries from tracelessness. (b)~Full, symmetric, and independent component counts as a function of tensor rank. (c)~Additions and multiplications for distinct couplings up to rank $5$, ordered by $J=l_1+l_2+l_3$, before compiler transformations. Independent components reduce Cartesian operation counts to values closer to sparse Clebsch--Gordan coupling. (d)~Lines of generated source code for the forward convolution as a function of maximum tensor rank.}
  \label{fig:component-cost}
\end{figure}

\textbf{Symbolic construction.} The generator averages over all $l_3!$ permutations, while the braces in Equation~\ref{eq:tp} sum over distinct placements. Let $\bm Q=\bm x_{l_1}\cdot n\cdot\bm y_{l_2}$ be the contracted product before trace removal. Its $l_1-n$ indices inherited from the first input and its $l_2-n$ indices inherited from the second input are separately symmetric, so every distinct placement occurs $(l_1-n)!\,(l_2-n)!$ times in the full permutation sum
\begin{equation*}
  \frac{1}{l_3!}\sum_{\pi\in S_{l_3}}\pi\bm Q
  =
  \frac{(l_1-n)!\,(l_2-n)!}{l_3!}\{\bm Q\}.
\end{equation*}
Here $S_{l_3}$ is the permutation group of the $l_3$ free indices, and $\pi\bm Q$ permutes these indices. Multiplication by $l_3!/[(l_1-n)!\,(l_2-n)!]=\binom{l_3}{l_1-n}$, where $(l_1-n)+(l_2-n)=l_3$, therefore converts the permutation average to the distinct-placement convention, and the generator multiplies each path by $\binom{l_3}{l_1-n}C_{l_1l_2l_3}$, where $C_{l_1l_2l_3}$ is the normalization constant in Equation~\ref{eq:tp}.

In Equation~\ref{eq:tp}, the $m>0$ terms remove traces from $\{\bm Q\}$, with coefficients $(-2)^m(2l_3-2m-1)!!/(2l_3-1)!!$ chosen so that the full sum is trace-free. The permutation average differs from $\{\bm Q\}$ only by a scalar factor, and removing traces preserves this relation. The generator therefore applies the factor derived above after trace removal, with no additional normalization.

\textbf{Symbolic derivatives.} For one path and feature channel, consider the weighted product $\bm c(w,\bm x,\bm y)=w\bm f(\bm x,\bm y)$, where $\bm f$ contains the independent output components. The scalar $w$ multiplies all $2l_3+1$ components. Given an incoming gradient $\bm g$ with respect to $\bm c$, the VJP computes the corresponding gradients $\bar{\bm x}$, $\bar{\bm y}$, and $\bar w$ with respect to the inputs and weight. Writing $a$ for an output-component index and $b_1,b_2$ for the two input-component indices gives
\begin{equation}
  \begin{aligned}
    \bar x_{b_1}
      &=w\sum_a g_a\frac{\partial f_a}{\partial x_{b_1}},
    &\bar y_{b_2}
      &=w\sum_a g_a\frac{\partial f_a}{\partial y_{b_2}},
    &\bar w
      &=\sum_a g_a f_a(\bm x,\bm y).
  \end{aligned}
  \label{eq:weighted-vjp}
\end{equation}
The weight gradient can also be written as $\bar w=\langle\bm g,\bm f(\bm x,\bm y)\rangle$, where $\langle\bm g,\bm z\rangle=\sum_a g_a z_a$ denotes the inner product over the independent output components.

To differentiate the VJP, we denote its unweighted input derivatives by $\operatorname{VJP}_1(\bm g,\bm y)$ and $\operatorname{VJP}_2(\bm g,\bm x)$, whose components are the sums in Equation~\ref{eq:weighted-vjp} without the factor $w$. Because $\bm f$ is bilinear, the first depends on $\bm y$ but not $\bm x$, and the second on $\bm x$ but not $\bm y$. Let $\bm v_x$, $\bm v_y$, and $v_w$ be the incoming gradients with respect to $\bar{\bm x}$, $\bar{\bm y}$, and $\bar w$. Differentiating the VJP gives gradients with respect to $\bm g$, $\bm x$, $\bm y$, and $w$, denoted by hats,
\begin{equation}
  \begin{aligned}
    \widehat{\bm g}
      &=w\bigl[\bm f(\bm v_x,\bm y)+\bm f(\bm x,\bm v_y)\bigr]
        +v_w\bm f(\bm x,\bm y),\\
    \widehat{\bm x}
      &=\operatorname{VJP}_1(\bm g,w\bm v_y+v_w\bm y),\\
    \widehat{\bm y}
      &=\operatorname{VJP}_2(\bm g,w\bm v_x+v_w\bm x),\\
    \widehat w
      &=\langle\bm g,
        \bm f(\bm v_x,\bm y)+\bm f(\bm x,\bm v_y)\rangle.
  \end{aligned}
  \label{eq:weighted-vjp-vjp}
\end{equation}
The contributions to $\widehat{\bm x}$ from differentiating $\bar{\bm y}$ and $\bar w$ combine into one evaluation of $\operatorname{VJP}_1$, since this map is linear in its second argument. The same applies to $\widehat{\bm y}$. No $\bm v_x$ term appears in $\widehat{\bm x}$ because $\bar{\bm x}$ is independent of $\bm x$, and likewise for $\widehat{\bm y}$. Finally, $\widehat w$ has no $v_w$ term because $\bar w$ is independent of $w$.

To use the generated kernels with PyTorch automatic differentiation, we register the explicit VJP as the derivative of the forward operation and the VJP-of-VJP as the derivative of the VJP. Force training also requires derivatives of the direction tensors described in Appendix~\ref{app:background-details}. These tensors are polynomial functions of the direction vector, so a separate generator differentiates their independent components symbolically to provide the corresponding VJP and VJP-of-VJP.

\textbf{Cost of the generated expressions.} Table~\ref{tab:operation-counts} compares the arithmetic required to evaluate the same set of couplings using full Cartesian components, independent Cartesian components, or sparse Clebsch--Gordan coefficients. For each maximum tensor rank, we sum the operation counts over all admissible paths, counting paths related by exchanging the inputs only once. Figure~\ref{fig:component-cost}(c) gives the counts for individual couplings. These counts exclude memory accesses and subsequent compiler transformations and therefore do not directly predict evaluation time.

\begin{table}[tbp]
  \centering
  \caption{Operation counts for Cartesian and spherical tensor products.}
  \label{tab:operation-counts}
  \begin{tabular}{crrrr}
    \toprule
    & & \multicolumn{2}{c}{Cartesian} & Spherical \\
    \cmidrule(lr){3-4}\cmidrule(lr){5-5}
    Maximum rank & Couplings & Full & Independent & Sparse CG \\
    \midrule
    3 & 15 & 1,299  & 615   & 609   \\
    4 & 27 & 13,802 & 2,774 & 2,141 \\
    5 & 42 & 79,128 & 7,505 & 4,951 \\
    \bottomrule
  \end{tabular}
\end{table}

The generated source grows with the maximum tensor rank (Figure~\ref{fig:component-cost}(d)). Symbolic generation takes seconds, while Triton compilation on first use took a few minutes at rank $4$ on our device, after which the compiled kernels are cached and reused.

\section{Executing Cartesian tensor products}
\label{app:schedule-details}

The kernels described in Section~\ref{sec:schedule} evaluate tensor products independently for each feature channel. When both inputs have $d$ feature channels, the kernel couples channel $k$ of one input with channel $k$ of the other. If one input has a single channel, as with direction tensors, it is reused for every channel of the other input. Separate linear maps perform dense mixing over channels and coupling paths. The same kernel implementation supports keeping coupling paths separate or summing their outputs, sharing weights, using symmetric inputs, and skipping unneeded derivative computations. These options are fixed at compile time.

\begin{figure}[t]
  \centering
  \includegraphics[width=0.98\textwidth]{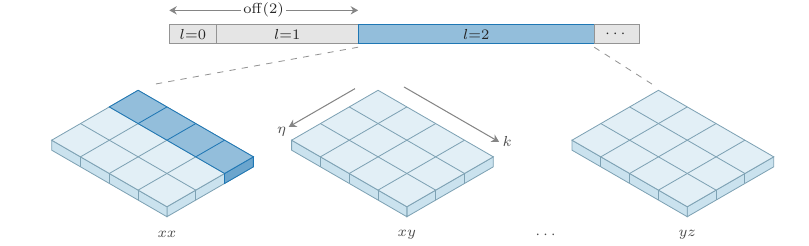}
  \caption{\textbf{Memory layout of Cartesian tensor features.} Each row $X_{r,:}$ stores tensor ranks in contiguous blocks, with $\operatorname{off}(l)$ giving the offset of the block for rank $l$ within the row. The expanded rank-$2$ block shows how each stored Cartesian component contains $P_2$ coupling paths and $d_X$ feature channels. The darker cells highlight the contiguous channels of one component--path pair.}
  \label{fig:layout-ownership}
\end{figure}

\textbf{Data layout.} An equivariant array $X$ has one row per node or edge and $d_X$ feature channels. A minibatch concatenates the rows of its graphs and adjusts the edge indices. This allows the kernels to process the batch without a separate batch dimension. Each row contains contiguous blocks in increasing tensor rank. Within the block for rank $l$, entries are ordered by the independent Cartesian component, then by the coupling path, and finally by the feature channel (Figure~\ref{fig:layout-ownership}). Here $P_l$ counts the paths stored separately for output rank $l$, with $P_l=1$ for node features and direction tensors. For component index $c\in\{0,\ldots,2l\}$, path index $\eta\in\{0,\ldots,P_l-1\}$, and channel index $k\in\{0,\ldots,d_X-1\}$, the entry's offset within the row is
\begin{equation*}
  \operatorname{off}(l)+(cP_l+\eta)d_X+k,
  \qquad
  \operatorname{off}(l)=\sum_{l'<l}(2l'+1)P_{l'}d_X.
\end{equation*}
Keeping ranks up to $L$ therefore selects the first $\operatorname{off}(L+1)$ entries of each row without copying or rearranging the features. The product layer uses this selection when intermediate ranks exceed the block’s output ranks, including when the last message-passing block returns only scalars.

\textbf{Node-local and edge-parallel kernels.} The shared implementation evaluates tensor products for a tile of nodes or edges. For node-local products, it reads the corresponding input rows and stores each coupling path's output separately. For edge convolutions, the edge indices identify the input rows and the receiver to which each contribution is added atomically. Summing coupling paths into a common output also uses atomic additions in this implementation. The receiver-sorted kernel instead sums both paths and incoming-edge contributions in local accumulators and stores each result once. Neither convolution kernel stores an intermediate tensor of edge messages.

Reusing an input or weight in the forward computation requires adding the corresponding gradient contributions during differentiation. When an input is reused across feature channels, its gradient sums the contributions from those channels. Similarly, when a weight is shared across rows, its gradient sums the contributions from those rows.

\textbf{Receiver-sorted forward.} Section~\ref{sec:schedule} indexes edges by $e$, while Section~\ref{sec:background} identifies them by the atom pair $uv$. All edges are sorted by receiver, and a row-pointer array identifies the incoming edges of receiver $u$ by the range $[\texttt{rowptr}[u],\texttt{rowptr}[u+1])$. The sender of edge $e$ is provided by $\texttt{sender}[e]$. Each coupling path $\eta=(l_1,l_2\to l)$ combines rank $l_1$ of the sender features with rank $l_2$ of the direction tensor $\bm y_e=\bm T(\hat{\bm r}_{u_ev_e})$ to produce output rank $l$. When both inputs have the same number of channels, we write the second input as $\bm y_{ek}$ and load it for each channel tile. Using these inputs and the precomputed scalar path weights, each program evaluates the tensor products over its receiver's incoming edges and stores the accumulated sums after processing all of them.

To limit the number of sums accumulated simultaneously, we group output ranks with a target of $100$ stored components per channel, counting the output components of each coupling path separately. Outputs of the same tensor rank are kept in one group, even if their combined component count exceeds the target. For the fused VJP-of-VJP, we reduce the target number of output components per group to limit register use. We launch a separate kernel for each group of output ranks $\mathcal G$. Within each launch, we divide the feature channels into tiles $\mathcal K$, and each program processes one receiver and one tile. It evaluates all coupling paths that produce ranks in $\mathcal G$, denoted by $\mathcal P_{\mathcal G}$ in Algorithm~\ref{alg:csr-fwd}. Each program stores the output components for its receiver and channel tile, without atomic additions. During code generation, the operations for every path are written out explicitly and applied independently to the channels processed by that program.

\begin{figure}[t]
\centering
\begin{minipage}[t]{0.49\textwidth}
\vspace{0pt}
\setlength{\intextsep}{0pt}
\begin{algorithm}[H]
\caption{Receiver-sorted VJP}
\label{alg:csr-vjp}
\fontsize{8.5}{11.05}\selectfont
\algrenewcommand\algorithmicindent{0.8em}
\begin{algorithmic}[1]
\Require Receiver-sorted arrays \texttt{rowptr}, \texttt{sender}
\Require Tensor inputs $\bm x,\bm y$ and scalar weights $w$
\Require Rank groups $\mathcal G$, paths $\mathcal P_{\mathcal G}$, and channel tiles $\mathcal K$
\Require Cotangent $\bm g$, chunk length $C$, and slots $S$
\State Initialize $\bar{\bm x},\bar{\bm y}\gets\bm0$
\For{each rank group $\mathcal G$}
  \ForAll{receivers $u$, slots $\sigma<S$, and tiles $\mathcal K$}
    \State Load $e_0,e_1\gets\texttt{rowptr}[u],\texttt{rowptr}[u+1]$
    \State Set $j\gets e_0+\sigma C$
    \If{$j\ge e_1$}
      \State \textbf{exit program}
    \EndIf
    \State Load $\bm g_{u\eta k}$ for $\eta\in\mathcal P_{\mathcal G}$, $k\in\mathcal K$
    \While{$j<e_1$}
      \For{$e=j,\ldots,\min(j+C,e_1)-1$}
        \State Load $v\gets\texttt{sender}[e]$, $\bm x_{v\mathcal K}$, and $\bm y_e$
        \State Initialize $\bm a^x_k\gets\bm0$ for $k\in\mathcal K$ and $\bm a^y\gets\bm0$
        \For{$\eta\in\mathcal P_{\mathcal G}$, $k\in\mathcal K$}
          \State Load $w_{e\eta k}$
          \State $\bm a^x_k\mathrel{+}=w_{e\eta k}\operatorname{VJP}_{1,\eta}(\bm g_{u\eta k},\bm y_e)$
          \State $\bm a^y\mathrel{+}=w_{e\eta k}\operatorname{VJP}_{2,\eta}(\bm g_{u\eta k},\bm x_{vk})$
          \State Store $\bar w_{e\eta k}\gets\langle\bm g_{u\eta k},\bm f_\eta(\bm x_{vk},\bm y_e)\rangle$
        \EndFor
        \State $\operatorname{AtomicAdd}(\bar{\bm x}_{v\mathcal K},\bm a^x_{\mathcal K})$
        \State $\operatorname{AtomicAdd}(\bar{\bm y}_e,\bm a^y)$
      \EndFor
      \State Set $j\gets j+SC$
    \EndWhile
  \EndFor
\EndFor
\end{algorithmic}
\end{algorithm}
\end{minipage}\hfill
\begin{minipage}[t]{0.49\textwidth}
\vspace{0pt}
\setlength{\intextsep}{0pt}
\begin{algorithm}[H]
\caption{Receiver-sorted VJP-of-VJP}
\label{alg:csr-vjp-vjp}
\fontsize{8.5}{10.2}\selectfont
\algrenewcommand\algorithmicindent{0.6em}
\begin{algorithmic}[1]
\Require Receiver-sorted arrays \texttt{rowptr}, \texttt{sender}
\Require Tensor inputs $\bm x,\bm y$ and scalar weights $w$
\Require Rank groups $\mathcal G$, paths $\mathcal P_{\mathcal G}$, and channel tiles $\mathcal K$
\Require Cotangents $\bm g$, $\bm v_x$, $\bm v_y$, $v_w$
\State Initialize $\widehat{\bm x},\widehat{\bm y}\gets\bm0$
\For{each rank group $\mathcal G$}
  \ForAll{receivers $u$ and channel tiles $\mathcal K$}
    \State Load $e_0,e_1\gets\texttt{rowptr}[u],\texttt{rowptr}[u+1]$
    \State Load $\bm g_{u\eta k}$ for $\eta\in\mathcal P_{\mathcal G}$, $k\in\mathcal K$
    \State Initialize $\widehat{\bm g}^{\mathrm{loc}}_{\eta k}\gets\bm0$ for $\eta\in\mathcal P_{\mathcal G}$, $k\in\mathcal K$
    \For{$e=e_0,\ldots,e_1-1$}
      \State Load $v\gets\texttt{sender}[e]$, $\bm x_{v\mathcal K}$, and $\bm y_e$
      \State Load $\bm v_{x,v\mathcal K}$, $\bm v_{y,e}$
      \State Initialize $\widehat{\bm a}^x_k\gets\bm0$ for $k\in\mathcal K$ and $\widehat{\bm a}^y\gets\bm0$
      \For{$\eta\in\mathcal P_{\mathcal G}$, $k\in\mathcal K$}
        \State Load $w_{e\eta k}$, $v_{w,e\eta k}$
        \State $\bm s_{e\eta k}\gets\bm f_\eta(\bm v_{x,vk},\bm y_e)+\bm f_\eta(\bm x_{vk},\bm v_{y,e})$
        \State $\begin{aligned}[t]
          \widehat{\bm g}^{\mathrm{loc}}_{\eta k}\mathrel{+}={}&w_{e\eta k}\bm s_{e\eta k}\\
          &+v_{w,e\eta k}\bm f_\eta(\bm x_{vk},\bm y_e)
        \end{aligned}$
        \State $\begin{aligned}[t]
          \widehat{\bm a}^x_k\mathrel{+}=\operatorname{VJP}_{1,\eta}\bigl(&\bm g_{u\eta k},\\
          &w_{e\eta k}\bm v_{y,e}+v_{w,e\eta k}\bm y_e\bigr)
        \end{aligned}$
        \State $\begin{aligned}[t]
          \widehat{\bm a}^y\mathrel{+}=\operatorname{VJP}_{2,\eta}\bigl(&\bm g_{u\eta k},\\
          &w_{e\eta k}\bm v_{x,vk}+v_{w,e\eta k}\bm x_{vk}\bigr)
        \end{aligned}$
        \State Store $\widehat w_{e\eta k}\gets\langle\bm g_{u\eta k},\bm s_{e\eta k}\rangle$
      \EndFor
      \State $\operatorname{AtomicAdd}(\widehat{\bm x}_{v\mathcal K},\widehat{\bm a}^x_{\mathcal K})$
      \State $\operatorname{AtomicAdd}(\widehat{\bm y}_e,\widehat{\bm a}^y)$
    \EndFor
    \State Store $\widehat{\bm g}_{u\eta k}\gets\widehat{\bm g}^{\mathrm{loc}}_{\eta k}$ for $\eta\in\mathcal P_{\mathcal G}$, $k\in\mathcal K$
  \EndFor
\EndFor
\end{algorithmic}
\end{algorithm}
\end{minipage}
\end{figure}

\textbf{Receiver-sorted derivatives.} Let $\bm g_{u\eta k}$ be the incoming gradient of $\bm z_{u\eta k}$, and let $\operatorname{VJP}_{1,\eta}$ and $\operatorname{VJP}_{2,\eta}$ denote the unweighted input VJPs defined in Appendix~\ref{app:lowering-details}. For each edge, the program sums the contributions from its assigned coupling paths before atomically adding the resulting sums to the input gradients, as shown in Algorithm~\ref{alg:csr-vjp}. Atomic additions combine contributions from different receivers to the gradient of the same sender features. They also combine contributions to the gradient of a direction tensor from programs processing different channel tiles or groups of output ranks. Each weight gradient belongs to one edge, path, and channel and is stored without atomic addition. The derivative kernels use the same indexing as the forward kernel. Edge index $e$ identifies the direction tensor, scalar weights, and their gradients, while sender index $v_e$ identifies the sender features and their gradients. The edges therefore remain sorted by receiver throughout differentiation.

To process receivers with many neighbors in parallel, we divide each receiver's incoming edges into chunks of at most $C$ edges. For each rank group and channel tile, $S$ programs share the work. Program $\sigma$ starts with chunk $\sigma$ and advances by $S$ chunks, processing $\sigma,\sigma+S,\ldots$. Programs with no assigned edges skip the computation. Each active program loads the receiver gradient once and reuses it for all its edges. We select the number of programs $S$ before launch based on the batch's average number of neighbors. Autotuning selects $C$ together with the other kernel settings, including a large-chunk option in which program zero processes all incoming edges when they fit in one chunk.

The forward output and $\widehat{\bm g}$ in the VJP-of-VJP both require sums over all incoming edges of a receiver. Dividing these edges among programs would require combining their partial sums. We therefore restrict edge chunking to the VJP. In Algorithm~\ref{alg:csr-vjp-vjp}, each program processes all incoming edges of one receiver for its assigned group of output ranks and channel tile. For each edge, it evaluates Equation~\ref{eq:weighted-vjp-vjp} using the cotangents $\bm v_x$ of the sender's features in channel $k$, $\bm v_y$ of the edge's direction tensor, and $v_w$ of the weight for that edge, coupling path, and channel. Each program computes $\widehat{\bm g}$, the gradient with respect to the incoming gradient $\bm g$, by summing contributions from all incoming edges of its receiver and stores the sum once. As in the VJP, programs atomically add contributions to shared input gradients $\widehat{\bm x}$ and $\widehat{\bm y}$ and store each weight gradient $\widehat w$ without atomic addition.

For maximum output ranks above four, we compute different derivative outputs in separate kernel launches to reduce the number of intermediate values held in registers. The VJP uses separate launches for $\bar{\bm x}$, $\bar{\bm y}$, and $\bar w$. For the VJP-of-VJP, one launch computes $\widehat{\bm g}$, another computes $\widehat{\bm x}$ and $\widehat{\bm y}$ together, and a third computes $\widehat w$. Each launch uses the generated component expressions and accumulates or stores its gradients as described above.

\section{Model architecture}
\label{app:architecture-details}

\begin{figure}[t]
  \centering
  \includegraphics[width=0.95\textwidth]{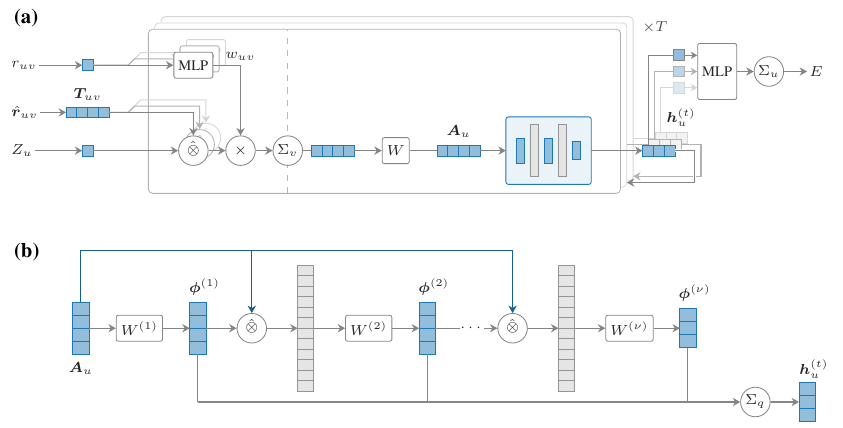}
  \caption{\textbf{Model architecture and fixed-width correlation ladder.} (a)~Each message-passing block combines a convolution with the product ladder. Blue bars show tensor features grouped by rank $l$, starting at $l=0$. Light grey lines show shared geometric inputs and blockwise scalar features combined in the readout to give $E$. The output of each block is also passed to the next block. (b)~The same atomic basis $\bm A_u$ is reused at every tensor product. Grey bars show the intermediate tensor product outputs with the retained coupling paths, compressed by $W^{(q)}$ to $d$ channels per retained rank, shown in blue. The output of each step enters the next product and also contributes to the final sum. Bar heights schematically indicate the number of tensor components. We skip residual updates, normalization, and nonlinear gates for simplicity.}
  \label{fig:architecture}
\end{figure}

Each atom is represented by a learned scalar embedding of its species. For each neighboring pair, the distance $r_{uv}$ determines the radial weights, while the unit direction $\hat{\bm r}_{uv}$ determines the Cartesian direction tensors used in the convolution. The features pass through $T$ message-passing blocks, each combining an equivariant convolution with the node-local product ladder of Section~\ref{sec:arch} (Figure~\ref{fig:architecture}). All blocks use the independent Cartesian components and memory layout described in Sections~\ref{sec:lowering}--\ref{sec:schedule}.

A residual update and equivariant RMS normalization follow both the convolution and the product ladder. Each residual update adds a learned linear map of the corresponding input to the output. The normalization computes scalar invariants with the generated $(l,l)\to0$ couplings and combines them across ranks and channels into one normalization scale per atom, without batch statistics~\citep{Liao2026}. A Euclidean norm of the stored components would not be rotationally invariant because the Cartesian coordinates are not orthonormal. After normalization, nonlinear functions of the scalar features gate the equivariant channels before the product ladder~\citep{Weiler2018}. A learned output map transforms the gated features, and its output is added to the input of the gated layer. Initializing this map to zero makes the layer an identity at initialization. The resulting features define the atomic basis $\bm A_u$ supplied to the product ladder.

After summing contributions through degree $\nu$, the product ladder is followed by a residual update and normalization to give the output of a message-passing block. Intermediate blocks retain ranks up to $l_{\max}^{\mathrm{prop}}$, while the final block outputs only scalars. Scalar channels from all $T$ blocks are concatenated and passed through a readout multi-layer perceptron (MLP) to obtain the atomic energy $E_u$. The atomic energies are scaled and shifted using fixed values determined from the training set. The per-element shifts are obtained by ridge-regularized least squares~\cite{Zaverkin2021}, while the output scale is set to the root-mean-square (RMS) force component. Finally, the total energy is obtained by summing the atomic energies, $E=\sum_u E_u$.

\textbf{Equivariant convolution.} For message-passing block $t$, $\bm h^{(t)}_{vkl_1}$ denotes channel $k$ of the rank-$l_1$ features at sender $v$, with $d$ channels per retained rank. Before the convolution, a $d\times d$ map mixes the channels independently at each rank. Denoting the resulting features by $\bar{\bm h}^{(t)}$, the convolution aggregates their contributions from neighbors $v\in\mathcal N(u)$ at receiver $u$,
\begin{equation}
  \bm A^{(t)}_{u\eta kl}
  =\frac{1}{\sqrt{\bar N}}
   \sum_{v\in\mathcal N(u)}
   w^{(t)}_{\eta k}(r_{uv})
   \left(
     \bar{\bm h}^{(t)}_{vkl_1}
     \hat{\otimes}_{\mathrm{Cart}}
     \bm T_{l_2}(\hat{\bm r}_{uv})
   \right)_l,
  \label{eq:abasis}
\end{equation}
where $\eta=(l_1,l_2\to l)$ is an admissible coupling path. The factor $1/\sqrt{\bar N}$ uses the mean number of neighbors in the training data. The receiver-sorted kernel of Section~\ref{sec:schedule} evaluates Equation~\ref{eq:abasis}, retaining the contribution of each path until a rank-wise linear map mixes the $P_l$ paths and $d$ channels into $d$ output channels. For $l_{\max}^{\mathrm{edge}}=l_{\max}^{\mathrm{prod}}=3$ and $l_{\max}^{\mathrm{prop}}=2$, the first convolution has four paths because the sender features are scalar, while subsequent convolutions have 17.

The radial weights $w^{(t)}_{\eta k}(r_{uv})$ are produced by a bias-free MLP with one output per coupling path and channel. Its input is a Chebyshev expansion of the distance multiplied by a polynomial cutoff envelope whose value and first two derivatives vanish at $r_{\max}$~\citep{Gasteiger2020}. With a smooth activation satisfying $\sigma(0)=0$, the radial weights and their first two derivatives also vanish at the cutoff. Edges can therefore enter or leave the cutoff without introducing discontinuities in the energy or its first two derivatives.

\textbf{Factorized fixed-width correlations.} A general bilinear equivariant map assigns a coefficient $\widetilde W^{(q)}_{l,k(\eta k_1k_2)}$ to each coupling path $\eta$, pair of input channels $(k_1,k_2)$, and output channel $k$. The product in Equation~\ref{eq:ladder} couples only matching input channels, giving
\begin{equation*}
  \widetilde W^{(q)}_{l,k(\eta k_1k_2)}
  =\delta_{k_1k_2}\,W^{(q)}_{l,k(\eta k_1)}.
\end{equation*}
The coefficients therefore vanish when $k_1\ne k_2$, while $W^{(q)}$ mixes all retained paths and channels densely within each output rank. The separate path contributions in $\bm B^{(q)}$ are stored until this map produces $\bm\phi^{(q)}$. When both inputs and the output retain tensor ranks $l=0\dots3$, $\bm B^{(q)}$ contains $99d$ independent Cartesian components across its coupling paths. Mixing reduces this to $16d$ components, corresponding to $d$ channels at each retained rank (Figure~\ref{fig:architecture}(b)).

\textbf{Tensor-train interpretation.} Expanding the recurrence in Equation~\ref{eq:ladder} expresses $\bm\phi^{(\nu)}$ as a polynomial in the atomic basis. For a fixed output rank $l$, let $\bm\eta=(\eta_2,\ldots,\eta_\nu)$ denote a retained sequence of coupling paths whose final product has rank $l$. We write $i_1,\ldots,i_\nu$ for the channels of the successive basis factors and $j$ for the output channel. Suppressing the tensor ranks specified by $\bm\eta$, we obtain
\begin{equation}
  \bm\phi^{(\nu)}_{uj}
  =\sum_{\bm\eta}\sum_{i_1,\ldots,i_\nu}
   \Theta^j_{i_1\cdots i_\nu}(\bm\eta)\,
   \bigl[\bm A_{ui_1}\hat{\otimes}_{\mathrm{Cart}}\cdots
     \hat{\otimes}_{\mathrm{Cart}}\bm A_{ui_\nu}\bigr]_{\bm\eta}.
  \label{eq:ladder-unrolled}
\end{equation}
The products in brackets are evaluated from left to right along the paths in $\bm\eta$. For each retained path sequence, the coefficient tensor factors into the learned maps of the successive ladder steps. This gives the tensor-train form~\citep{Oseledets2011}
\begin{equation}
  \Theta^j_{i_1\cdots i_\nu}(\bm\eta)
    =\sum_{k_1,\ldots,k_{\nu-1}}
      W^{(1)}_{k_1i_1}
      \prod_{q=2}^{\nu}G^{(q)}_{k_{q-1},i_q,k_q}(\eta_q),
      \qquad k_\nu=j,
\end{equation}
with
\begin{equation}
  G^{(q)}_{k_{q-1},i_q,k_q}(\eta_q)
      =\delta_{k_{q-1}i_q}\,W^{(q)}_{k_q(\eta_q i_q)}.
  \label{eq:tt-core}
\end{equation}
Here $W^{(1)}$ mixes the channels of the first basis factor, while each $G^{(q)}$ describes one subsequent product step. The indices $k_q$ connect successive steps, and each takes $d$ values, set by the model width. The Kronecker delta pairs each incoming channel with the same channel of the atomic basis, while $W^{(q)}$ mixes the resulting path and channel contributions into the outgoing channels. The recurrence uses this factorization implicitly and evaluates it directly without constructing $\Theta$. Contributions from earlier ladder steps follow the same factorization truncated after $q$ basis factors, and the ladder output sums contributions of degrees $1,\ldots,\nu$.

\textbf{Path-reduced products.} For the path-reduced products, each path and channel has a learned weight $a^{(q)}_{\eta k}=1+\delta^{(q)}_{\eta k}$, shared across atoms and initialized with $\delta^{(q)}_{\eta k}=0$. At the first step, there is one weight per rank and channel, $\bm\phi^{(1)}_{ukl}=a^{(1)}_{lk}\bm A_{ukl}$. For $q=2,\ldots,\nu$, the recurrence becomes
\begin{equation*}
  \bm\phi^{(q)}_{ukl}
    =\frac{1}{\sqrt{P_{q,l}}}
      \sum_{\eta=(l_1,l_2\to l)}
      a^{(q)}_{\eta k}
      \left(\bm\phi^{(q-1)}_{ukl_1}
        \hat{\otimes}_{\mathrm{Cart}}\bm A_{ukl_2}\right)_l,
    \qquad l\le l_q,
\end{equation*}
where $P_{q,l}$ is the number of coupling paths producing rank $l$. The kernel accumulates these paths as it evaluates the products, without mixing channels. The step outputs are then combined and passed through a rank-wise linear map $W$ that mixes the channels to produce
\begin{equation*}
  \bm z_{ukl}
    =\sum_{k'} W_{l,kk'}\left(\sum_{q=1}^{\nu}\bm\phi^{(q)}_{uk'l}\right),
    \qquad l\le l_{\max}^{\mathrm{out}}.
\end{equation*}

\section{Experimental details}
\label{app:data-training}
\label{app:protocol}

\textbf{Datasets and splits.} SPICE-MACE-OFF is a neutral-molecule dataset based on SPICE~\citep{Eastman2023} that was originally used to train and evaluate MACE-OFF23~\citep{Kovacs2025}. It covers H, C, N, O, F, P, S, Cl, Br, and I, with $951{,}005$ structures in the training set and $50{,}195$ in the test set. We train three models, each using a randomly selected validation set of $47{,}550$ structures and the remaining $903{,}455$ structures for training. The test set comprises seven subsets: PubChem, DES370K monomers and dimers, dipeptides, solvated amino acids, water, and QMugs.

MAD~\citep{Mazitov2025} contains $95{,}595$ structures spanning $85$ elements, including bulk and two-dimensional crystals, molecular crystals and fragments, surfaces, clusters, and structures with distorted geometries or randomized chemical compositions. We use the released split of $76{,}483$ training, $9{,}566$ validation, and $9{,}546$ test structures for all three runs, varying model initialization. We further evaluate performance across materials and molecular systems using the PET-MAD benchmark~\citep{Mazitov2025b}. Its $1{,}562$ structures comprise $360$ MAD~\citep{Mazitov2025}, $136$ MPtrj~\citep{Deng2023}, $555$ MatBench~\citep{Riebesell2025}, $200$ Alexandria~\citep{Schmidt2023}, $78$ OC2020-S2EF~\citep{Chanussot2021}, $99$ SPICE~\citep{Eastman2023}, and $134$ MD22~\citep{Chmiela2023} structures.

\textbf{Model architecture and training.} All FlashCart models use eight Chebyshev radial functions evaluated at cosine-transformed interatomic distances and a polynomial cutoff envelope with $p=6$~\citep{Gasteiger2020}. The radial and readout MLPs each have two hidden layers of width $64$. Within each message-passing block, we apply equivariant RMS normalization and scalar SiLU gates as described in Appendix~\ref{app:architecture-details}. The initial embedding depends on the chemical species, while the convolution and product weights are shared across all species. The architectural comparisons in Figures~\ref{fig:efficiency} and~\ref{fig:ablation} examine changes in correlation order, channel width, message-passing depth, and maximum tensor ranks. Table~\ref{tab:model-configurations} summarizes these configurations, the five FlashCart models in Figure~\ref{fig:pareto}, and the model trained on MAD. We also compare correlation orders at approximately fixed model size by reducing channel width as order increases, keeping the parameter count between $0.913$M and $0.945$M. This experiment uses configurations with two message-passing blocks, maximum propagated rank $1$, and maximum edge and product ranks from $1$ to $4$.

\begin{table}[t]
\centering
\small
\setlength{\tabcolsep}{4pt}
\caption{\textbf{Model configurations.} $T$, $\nu$, and $d$ denote the number of message-passing blocks, correlation order, and channel width, respectively, while $l_{\max}^{\mathrm{edge}}=l_{\max}^{\mathrm{prod}}$ gives the shared maximum rank of the edge and product tensors. For the models in Figures~\ref{fig:efficiency} and~\ref{fig:ablation}, default values apply to parameters that are held fixed. The labels ``Ladder'' and ``Reduced'' denote the correlation ladder and path-reduced products described in Appendix~\ref{app:architecture-details}.}
\label{tab:model-configurations}
\begin{tabular}{lccccc l}
\toprule
Configuration & $T$ & $\nu$ & $d$ & $l_{\max}^{\mathrm{prop}}$ & $l_{\max}^{\mathrm{edge}}=l_{\max}^{\mathrm{prod}}$ & Product \\
\midrule
\multicolumn{7}{c}{Models shown in Figures~\ref{fig:efficiency} and~\ref{fig:ablation}} \\
\midrule
Default values & 2 & 2 & 64 & 1 & 3 & Ladder \\
Explored values & $1$--$4$ & $2$--$5$ & $32,64,96,128$ & $0$--$3$ & $1$--$4$ & Ladder \\
\midrule
\multicolumn{7}{c}{Models shown in Figure~\ref{fig:pareto} and Table~\ref{tab:pareto-data}} \\
\midrule
FlashCart-69k & 2 & 3 & 20 & 2 & 3 & Reduced \\
FlashCart-0.5M & 3 & 5 & 32 & 2 & 3 & Ladder \\
FlashCart-1.1M & 3 & 3 & 64 & 2 & 3 & Ladder \\
FlashCart-3.0M & 3 & 5 & 88 & 2 & 3 & Ladder \\
FlashCart-5.6M & 4 & 5 & 104 & 2 & 3 & Ladder \\
\midrule
\multicolumn{7}{c}{Models shown in Tables~\ref{tab:mad} and~\ref{tab:mad-test}} \\
\midrule
FlashCart-2.3M & 3 & 3 & 96 & 2 & 3 & Ladder \\
\bottomrule
\end{tabular}
\end{table}

\begin{table}[t]
\centering
\caption{\textbf{Training parameters.} We use the target batch size and training-set averages to determine how many atoms and edges each batch can contain, so the number of structures per batch can vary. For distributed training, we divide the target batch size across GPUs. EMA denotes the exponential moving average of model parameters.}
\label{tab:training-settings}
\begin{tabular}{lccc}
\toprule
Hyperparameter & \shortstack{Figures~\ref{fig:efficiency} and~\ref{fig:ablation}} & \shortstack{Figure~\ref{fig:pareto} and Table~\ref{tab:pareto-data}} & \shortstack{Tables~\ref{tab:mad} and~\ref{tab:mad-test}} \\
\midrule
Cutoff radius & 6.0 & 6.0 & 5.0 \\
Maximum epochs & 150 & 400 & 400 \\
Warmup epochs & 20 & 20 & 20 \\
Cosine-decay epochs & 130 & 380 & 380 \\
Target batch size & 128 & 128 & 192 \\
Number of random seeds & 3 & 3 & 3 \\
Peak learning rate & $5\times10^{-3}$ & $5\times10^{-3}$ & $2.5\times10^{-3}$ \\
Weight decay & $3\times10^{-4}$ & $3\times10^{-4}$ & $1.2\times10^{-2}$ \\
EMA decay & 0.999 & 0.999 & 0.999 \\
Force weight & 1.0 & 1.0 & 0.1 \\
\bottomrule
\end{tabular}
\end{table}

Before training, we determine the per-element energy shifts, output scale, and convolution normalization, which remain fixed throughout training. We fit the energy shifts by ridge regression~\citep{Zaverkin2021}, with isolated-atom energies as reference values for SPICE-MACE-OFF. The output scale is set to the root mean square of the force components and the convolution normalization to the average number of neighbors per atom in the training set. We use single precision and the training hyperparameters listed in Table~\ref{tab:training-settings}, with Muon for matrix weights and AdamW for the remaining parameters, excluding embeddings and biases from weight decay. The learning rate increases linearly from $10\%$ of its peak value during warmup, then follows a cosine decay to $1\%$ of the peak. For a batch with $B$ structures and $N=\sum_b N_b$ atoms, the training loss reads
\begin{equation}
  \mathcal L=\frac{1}{B}\sum_{b=1}^B
    \frac{|E_b-E_b^{\mathrm{ref}}|}{N_b}
    +w_F\,\frac{1}{N}\sum_{i=1}^{N}
      \|\bm F_i-\bm F_i^{\mathrm{ref}}\|_2,
  \label{eq:training-loss}
\end{equation}
where $E_b$ and $\bm F_i$ are the predicted energies and forces, while $E_b^\mathrm{ref}$ and $\bm F_i^\mathrm{ref}$ denote the corresponding reference values. Energies and forces are given in eV and eV/\AA, and Table~\ref{tab:training-settings} provides the force weight $w_F$ in \AA. We evaluate the model using an exponential moving average of its parameters and select the epoch with the lowest sum of validation energy MAE (eV/atom) and force MAE (eV/\AA).

\textbf{Normalized error.} We combine energy and force MAEs into normalized error (NE), following Ref.~\citenum{Gubaev2023},
\begin{equation}
 \text{NE} = \frac{1}{2} \left(\frac{\text{F-MAE}}{\text{max}\,\text{F-MAE}} + \frac{\text{E-MAE}}{\text{max}\,\text{E-MAE}}\right).
 \label{eq:normalized-error}
\end{equation}
Here E-MAE and F-MAE are the energy and force MAEs, and the denominators are their respective maxima over the models included in each figure. For Figure~\ref{fig:pareto}, the MAEs are averaged over the seven SPICE-MACE-OFF test subsets reported in Table~\ref{tab:pareto-data}. For Figure~\ref{fig:efficiency}, the MAEs are evaluated over the entire test set.

\textbf{Runtime and memory measurements.} We measure runtime and memory on an NVIDIA RTX PRO 6000 Blackwell Server Edition using PyTorch 2.11.0 and CUDA 12.8. All computations use single precision with TF32 disabled. We use periodic diamond cells with lattice constant $3.567$~\AA\ and precompute neighbor lists using each model's cutoff radius. Energy and force evaluation uses $1000$ atoms for the models in Figure~\ref{fig:pareto} and Tables~\ref{tab:mad}, \ref{tab:pareto-data}, and~\ref{tab:mad-test}, and $4096$ atoms for those in Figures~\ref{fig:efficiency} and~\ref{fig:ablation}. We also vary the system size from $216$ to $4096$ atoms to examine runtime and memory scaling in Figure~\ref{fig:pareto}. For Figure~\ref{fig:backend}, we use $1000$ atoms for inference and $512$ for training because the uncompiled PyTorch implementation runs out of GPU memory when training on $1000$ atoms. Training timings additionally include the loss and its parameter gradients, but exclude optimizer updates and data loading.

For each SPICE-MACE-OFF model benchmarked on our GPUs, we select the implementation with the lowest measured runtime. We enable compilation for FlashCart, MACE, PET-S, and DPA4. FlashCart uses generated Triton kernels, MACE uses cuEquivariance~\citep{Geiger2024}, and DPA4 uses DeePMD-kit's Triton kernels~\citep{Zeng2025}, while PET-L runs with TorchScript. For MAD, FlashCart also uses compilation, while PET-MAD runs in PyTorch without it. We estimate the eSEN runtimes from published results for an NVIDIA A100 GPU (80 GB)~\citep{Fu2025}.

\textbf{Kernel microbenchmarks.} We benchmark individual kernels using the same hardware and numerical precision as for the model benchmarks above, comparing FlashCart with cuEquivariance 0.11.0 and OpenEquivariance 0.6.8. For edge convolutions, we use OpenEquivariance's fused tensor-product convolution with deterministic evaluation disabled and cuEquivariance's channel-wise tensor product with fused gather and scatter operations. For node-local correlations, we compare FlashCart's path-reduced products and correlation ladder with cuEquivariance's symmetric contraction. The edge-convolution benchmark covers $30$ configurations with $1024$ nodes, $64$ or $128$ neighbors per node, $64$, $128$, or $256$ channels, and maximum tensor ranks from $1$ to $5$. Figure~\ref{fig:kernel-results}(a) shows the cases with $128$ neighbors and $128$ channels. To obtain the average runtime advantages reported in the main text, we divide each baseline's runtime by FlashCart's for each configuration and average these ratios over all $30$ configurations. The node-local benchmark uses $4096$ nodes, $64$, $128$, or $256$ channels, maximum tensor ranks from $1$ to $5$, and correlation orders from $2$ to $6$. Figure~\ref{fig:kernel-results}(b) shows the configurations with $128$ channels and maximum tensor ranks $3$ and $4$.

We measure the runtime of the forward, VJP, and VJP-of-VJP separately, excluding the preceding forward pass for the VJP and both the forward pass and first VJP for the VJP-of-VJP. Memory is measured as the peak increase in GPU allocation during each operation, excluding memory already used by its inputs and tensors saved from preceding computations. We report runtime and memory averaged over three independent runs. The VJP benchmark computes gradients with respect to the inputs and weights. The VJP-of-VJP benchmark differentiates the input gradients with respect to the inputs, weights, and incoming output gradient, following the derivative sequence needed for force training. It excludes contributions from differentiating the first VJP's weight gradients, corresponding to a zero weight-gradient cotangent in Equation~\ref{eq:weighted-vjp-vjp}.

\section{Results and discussion}
\label{app:additional-results}

\textbf{Kernel microbenchmarks.} The grey crosses in Figure~\ref{fig:kernel-results}(b) show cases in which we could not construct the cuEquivariance contraction or launch its kernel. For example, at rank $4$, order $4$, CPU RAM usage reached $767$~GiB before construction failed after approximately $2.7$ hours under the $1000$~GiB memory limit we set. At rank $3$, order $4$, we measured forward and VJP runtimes, but the VJP-of-VJP kernel failed to launch with a CUDA out-of-memory error on our NVIDIA RTX PRO 6000 Blackwell GPU (96 GB). Preparing the first VJP call also took about $20$ hours on the CPU at rank $4$, order $3$, and $37$ hours at rank $3$, order $4$.

\begin{figure}[tbp]
\centering
\includegraphics[width=\textwidth]{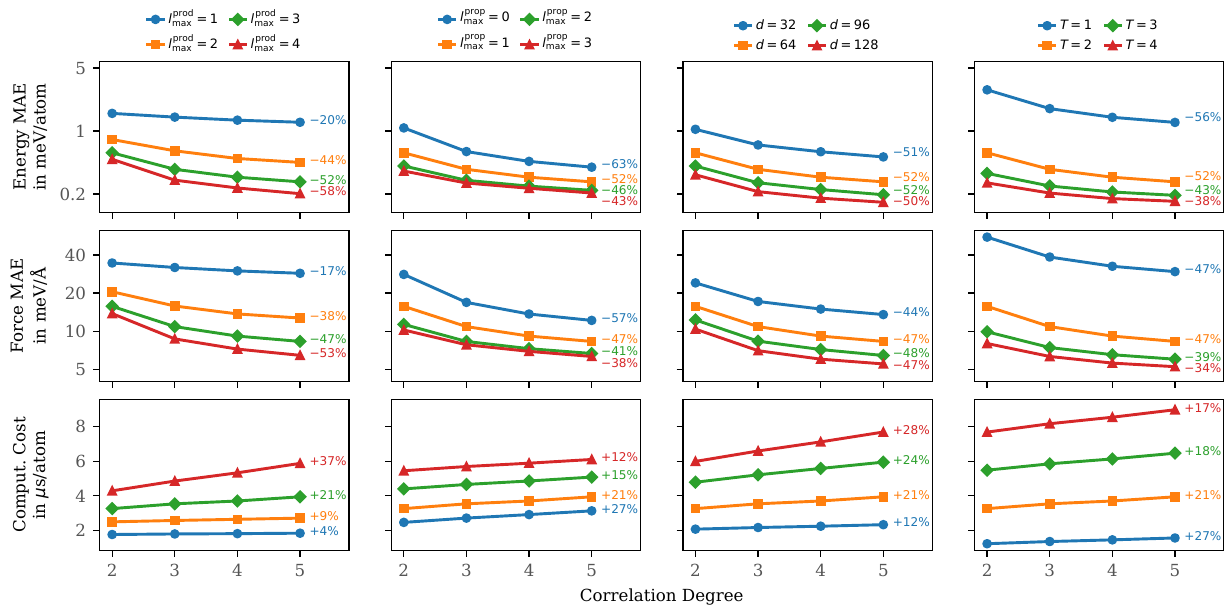}
\caption{\textbf{Effect of correlation order across model architectures.} Results for the models trained on SPICE-MACE-OFF and shown in Figure~\ref{fig:efficiency} are presented as energy MAE (top), force MAE (middle), and runtime (bottom) against correlation order $\nu$. From left to right, the columns show results for different maximum edge and product ranks (varied together), maximum propagated ranks, channel widths $d$, and numbers of message-passing blocks $T$. Unless varied in a column, models use maximum edge and product ranks of $l_\mathrm{max}^\mathrm{prod}=l_\mathrm{max}^\mathrm{edge}=3$, maximum propagated rank $l_\mathrm{max}^\mathrm{prop}=1$, $d=64$, and $T=2$. Percentages show the relative changes in error or runtime as $\nu$ increases from $2$ to $5$. Prediction errors are averaged over three independent training runs. Shaded bands show one standard deviation but are not visible at the plotted line width. Runtimes are measured for energy and force evaluation on $4096$ diamond atoms using one trained model per configuration.}
\label{fig:ablation}
\end{figure}

\textbf{Efficiency of higher-order correlations.} The $52$ configurations in Figure~\ref{fig:efficiency} cover correlation orders $\nu=2$ to $5$ for $13$ otherwise fixed architectures. These same configurations also let us compare channel widths, numbers of message-passing blocks, maximum propagated ranks, and jointly varied edge and product tensor ranks at each fixed correlation order, keeping the remaining architectural choices unchanged. For each comparison, we fit $\log\mathrm{NE}=a+b\log x$ by ordinary least squares, where $\mathrm{NE}$ is normalized error and $x$ is runtime or peak GPU memory per atom. The insets in Figure~\ref{fig:efficiency} show how normalized error changes with runtime or peak GPU memory when correlation order, channel width, the number of message-passing blocks, or tensor rank is varied. The fitted slopes quantify the relative change in error for a relative change in runtime or memory. Figure~\ref{fig:ablation} presents the same model results as Figure~\ref{fig:efficiency}, showing energy errors, force errors, and runtimes separately as correlation order increases. Its first column compares models with two message-passing blocks, $64$ channels, and maximum propagated rank $1$ at four fixed maximum edge and product ranks. Using the errors and runtimes in this column, we fit normalized error against runtime on logarithmic axes across correlation orders $\nu=2$ to $5$. The resulting slopes are $-5.0$, $-6.5$, $-3.7$, and $-2.6$ for maximum edge and product ranks $1$, $2$, $3$, and $4$, respectively.

\begin{table}[tbp]
\centering
\footnotesize
\setlength{\tabcolsep}{2.5pt}
\caption{\textbf{Energy and force errors on SPICE-MACE-OFF.} Results are shown separately for the seven test subsets for the models in Figure~\ref{fig:pareto}. Rows give energy (E) MAEs in meV/atom and force (F) MAEs in meV/\AA. We also report the arithmetic mean of the seven subset MAEs, separately for energy and forces. Normalized error (NE) combines these means as defined in Appendix~\ref{app:protocol}. FlashCart results show the mean and standard deviation across three independent training runs. We measure runtime and peak GPU memory for energy and force evaluation on a periodic diamond cell containing $1000$ atoms using an NVIDIA RTX PRO 6000 Blackwell GPU (96 GB). The lowest energy and force MAEs and normalized errors are shown in bold, taking the reported standard deviations into account.}
\label{tab:pareto-data}
\resizebox{\textwidth}{!}{%
\begin{tabular}{llr*{8}{r@{$\pm$}l}rrr}
\toprule
Model & & Params. & \multicolumn{2}{c}{PubChem} & \multicolumn{2}{c}{DES Mon.} & \multicolumn{2}{c}{DES Dim.} & \multicolumn{2}{c}{Dipept.} & \multicolumn{2}{c}{Solv. AA} & \multicolumn{2}{c}{Water} & \multicolumn{2}{c}{QMugs} & \multicolumn{2}{c}{Mean} & NE & $\mu$s/atom & MiB/atom \\
\cmidrule(r){1-3}\cmidrule(lr){4-19}\cmidrule(l){20-22}
MACE-OFF23-S~\citep{Kovacs2025} & E & 694k & \multicolumn{1}{r@{\phantom{$\pm$}}}{1.41} & & \multicolumn{1}{r@{\phantom{$\pm$}}}{1.04} & & \multicolumn{1}{r@{\phantom{$\pm$}}}{0.98} & & \multicolumn{1}{r@{\phantom{$\pm$}}}{0.84} & & \multicolumn{1}{r@{\phantom{$\pm$}}}{1.60} & & \multicolumn{1}{r@{\phantom{$\pm$}}}{1.67} & & \multicolumn{1}{r@{\phantom{$\pm$}}}{1.03} & & \multicolumn{1}{r@{\phantom{$\pm$}}}{1.224} & & \multirow{2}{*}{1.000} & \multirow{2}{*}{3.6} & \multirow{2}{*}{0.83} \\
 & F &  & \multicolumn{1}{r@{\phantom{$\pm$}}}{35.68} & & \multicolumn{1}{r@{\phantom{$\pm$}}}{17.63} & & \multicolumn{1}{r@{\phantom{$\pm$}}}{16.31} & & \multicolumn{1}{r@{\phantom{$\pm$}}}{25.07} & & \multicolumn{1}{r@{\phantom{$\pm$}}}{38.56} & & \multicolumn{1}{r@{\phantom{$\pm$}}}{28.53} & & \multicolumn{1}{r@{\phantom{$\pm$}}}{41.45} & & \multicolumn{1}{r@{\phantom{$\pm$}}}{29.03} & &  &  &  \\
MACE-OFF23-M~\citep{Kovacs2025} & E & 1.4M & \multicolumn{1}{r@{\phantom{$\pm$}}}{0.91} & & \multicolumn{1}{r@{\phantom{$\pm$}}}{0.63} & & \multicolumn{1}{r@{\phantom{$\pm$}}}{0.58} & & \multicolumn{1}{r@{\phantom{$\pm$}}}{0.52} & & \multicolumn{1}{r@{\phantom{$\pm$}}}{1.21} & & \multicolumn{1}{r@{\phantom{$\pm$}}}{0.76} & & \multicolumn{1}{r@{\phantom{$\pm$}}}{0.69} & & \multicolumn{1}{r@{\phantom{$\pm$}}}{0.757} & & \multirow{2}{*}{0.593} & \multirow{2}{*}{6.7} & \multirow{2}{*}{1.48} \\
 & F &  & \multicolumn{1}{r@{\phantom{$\pm$}}}{20.57} & & \multicolumn{1}{r@{\phantom{$\pm$}}}{9.36} & & \multicolumn{1}{r@{\phantom{$\pm$}}}{9.02} & & \multicolumn{1}{r@{\phantom{$\pm$}}}{14.27} & & \multicolumn{1}{r@{\phantom{$\pm$}}}{23.26} & & \multicolumn{1}{r@{\phantom{$\pm$}}}{15.27} & & \multicolumn{1}{r@{\phantom{$\pm$}}}{23.58} & & \multicolumn{1}{r@{\phantom{$\pm$}}}{16.48} & &  &  &  \\
MACE-OFF23-L~\citep{Kovacs2025} & E & 4.7M & \multicolumn{1}{r@{\phantom{$\pm$}}}{0.88} & & \multicolumn{1}{r@{\phantom{$\pm$}}}{0.59} & & \multicolumn{1}{r@{\phantom{$\pm$}}}{0.54} & & \multicolumn{1}{r@{\phantom{$\pm$}}}{0.42} & & \multicolumn{1}{r@{\phantom{$\pm$}}}{0.98} & & \multicolumn{1}{r@{\phantom{$\pm$}}}{0.83} & & \multicolumn{1}{r@{\phantom{$\pm$}}}{0.45} & & \multicolumn{1}{r@{\phantom{$\pm$}}}{0.670} & & \multirow{2}{*}{0.490} & \multirow{2}{*}{14.9} & \multirow{2}{*}{3.79} \\
 & F &  & \multicolumn{1}{r@{\phantom{$\pm$}}}{14.75} & & \multicolumn{1}{r@{\phantom{$\pm$}}}{6.58} & & \multicolumn{1}{r@{\phantom{$\pm$}}}{6.62} & & \multicolumn{1}{r@{\phantom{$\pm$}}}{10.19} & & \multicolumn{1}{r@{\phantom{$\pm$}}}{19.43} & & \multicolumn{1}{r@{\phantom{$\pm$}}}{13.57} & & \multicolumn{1}{r@{\phantom{$\pm$}}}{16.93} & & \multicolumn{1}{r@{\phantom{$\pm$}}}{12.58} & &  &  &  \\
\cmidrule(r){1-3}\cmidrule(lr){4-19}\cmidrule(l){20-22}
eSEN-3.2M~\citep{Fu2025} & E & 3.2M & \multicolumn{1}{r@{\phantom{$\pm$}}}{0.22} & & \multicolumn{1}{r@{\phantom{$\pm$}}}{0.17} & & \multicolumn{1}{r@{\phantom{$\pm$}}}{0.20} & & \multicolumn{1}{r@{\phantom{$\pm$}}}{0.10} & & \multicolumn{1}{r@{\phantom{$\pm$}}}{0.30} & & \multicolumn{1}{r@{\phantom{$\pm$}}}{0.24} & & \multicolumn{1}{r@{\phantom{$\pm$}}}{0.16} & & \multicolumn{1}{r@{\phantom{$\pm$}}}{0.199} & & \multirow{2}{*}{0.153} & \multirow{2}{*}{393$^\dagger$} & \multirow{2}{*}{--} \\
 & F &  & \multicolumn{1}{r@{\phantom{$\pm$}}}{6.10} & & \multicolumn{1}{r@{\phantom{$\pm$}}}{1.85} & & \multicolumn{1}{r@{\phantom{$\pm$}}}{2.77} & & \multicolumn{1}{r@{\phantom{$\pm$}}}{3.04} & & \multicolumn{1}{r@{\phantom{$\pm$}}}{5.76} & & \multicolumn{1}{r@{\phantom{$\pm$}}}{3.88} & & \multicolumn{1}{r@{\phantom{$\pm$}}}{5.70} & & \multicolumn{1}{r@{\phantom{$\pm$}}}{4.16} & &  &  &  \\
eSEN-6.5M~\citep{Fu2025} & E & 6.5M & \multicolumn{1}{r@{\phantom{$\pm$}}}{0.15} & & \multicolumn{1}{r@{\phantom{$\pm$}}}{0.13} & & \multicolumn{1}{r@{\phantom{$\pm$}}}{0.15} & & \multicolumn{1}{r@{\phantom{$\pm$}}}{0.07} & & \multicolumn{1}{r@{\phantom{$\pm$}}}{0.25} & & \multicolumn{1}{r@{\phantom{$\pm$}}}{0.15} & & \multicolumn{1}{r@{\phantom{$\pm$}}}{0.12} & & \multicolumn{1}{r@{\phantom{$\pm$}}}{0.146} & & \multirow{2}{*}{0.108} & \multirow{2}{*}{785$^\dagger$} & \multirow{2}{*}{--} \\
 & F &  & \multicolumn{1}{r@{\phantom{$\pm$}}}{4.21} & & \multicolumn{1}{r@{\phantom{$\pm$}}}{1.24} & & \multicolumn{1}{r@{\phantom{$\pm$}}}{2.12} & & \multicolumn{1}{r@{\phantom{$\pm$}}}{2.00} & & \multicolumn{1}{r@{\phantom{$\pm$}}}{3.68} & & \multicolumn{1}{r@{\phantom{$\pm$}}}{2.50} & & \multicolumn{1}{r@{\phantom{$\pm$}}}{3.78} & & \multicolumn{1}{r@{\phantom{$\pm$}}}{2.79} & &  &  &  \\
\cmidrule(r){1-3}\cmidrule(lr){4-19}\cmidrule(l){20-22}
DPA4-Neo~\citep{Li2026} & E & 1.1M & \multicolumn{1}{r@{\phantom{$\pm$}}}{0.24} & & \multicolumn{1}{r@{\phantom{$\pm$}}}{0.17} & & \multicolumn{1}{r@{\phantom{$\pm$}}}{0.18} & & \multicolumn{1}{r@{\phantom{$\pm$}}}{0.10} & & \multicolumn{1}{r@{\phantom{$\pm$}}}{0.27} & & \multicolumn{1}{r@{\phantom{$\pm$}}}{0.30} & & \multicolumn{1}{r@{\phantom{$\pm$}}}{0.12} & & \multicolumn{1}{r@{\phantom{$\pm$}}}{0.197} & & \multirow{2}{*}{0.154} & \multirow{2}{*}{39.2} & \multirow{2}{*}{6.44} \\
 & F &  & \multicolumn{1}{r@{\phantom{$\pm$}}}{5.92} & & \multicolumn{1}{r@{\phantom{$\pm$}}}{2.13} & & \multicolumn{1}{r@{\phantom{$\pm$}}}{2.16} & & \multicolumn{1}{r@{\phantom{$\pm$}}}{3.10} & & \multicolumn{1}{r@{\phantom{$\pm$}}}{6.69} & & \multicolumn{1}{r@{\phantom{$\pm$}}}{4.46} & & \multicolumn{1}{r@{\phantom{$\pm$}}}{5.36} & & \multicolumn{1}{r@{\phantom{$\pm$}}}{4.26} & &  &  &  \\
DPA4-Air~\citep{Li2026} & E & 5.1M & \multicolumn{1}{r@{\phantom{$\pm$}}}{0.13} & & \multicolumn{1}{r@{\phantom{$\pm$}}}{0.17} & & \multicolumn{1}{r@{\phantom{$\pm$}}}{0.11} & & \multicolumn{1}{r@{\phantom{$\pm$}}}{0.06} & & \multicolumn{1}{r@{\phantom{$\pm$}}}{0.13} & & \multicolumn{1}{r@{\phantom{$\pm$}}}{0.16} & & \multicolumn{1}{r@{\phantom{$\pm$}}}{0.07} & & \multicolumn{1}{r@{\phantom{$\pm$}}}{0.119} & & \multirow{2}{*}{0.089} & \multirow{2}{*}{78.2} & \multirow{2}{*}{8.14} \\
 & F &  & \multicolumn{1}{r@{\phantom{$\pm$}}}{3.59} & & \multicolumn{1}{r@{\phantom{$\pm$}}}{1.22} & & \multicolumn{1}{r@{\phantom{$\pm$}}}{1.23} & & \multicolumn{1}{r@{\phantom{$\pm$}}}{1.68} & & \multicolumn{1}{r@{\phantom{$\pm$}}}{3.59} & & \multicolumn{1}{r@{\phantom{$\pm$}}}{2.47} & & \multicolumn{1}{r@{\phantom{$\pm$}}}{2.91} & & \multicolumn{1}{r@{\phantom{$\pm$}}}{2.38} & &  &  &  \\
DPA4-Plus~\citep{Li2026} & E & 8.8M & \multicolumn{1}{r@{\phantom{$\pm$}}}{0.12} & & \multicolumn{1}{r@{\phantom{$\pm$}}}{0.12} & & \multicolumn{1}{r@{\phantom{$\pm$}}}{0.11} & & \multicolumn{1}{r@{\phantom{$\pm$}}}{\textbf{0.04}} & & \multicolumn{1}{r@{\phantom{$\pm$}}}{0.09} & & \multicolumn{1}{r@{\phantom{$\pm$}}}{0.11} & & \multicolumn{1}{r@{\phantom{$\pm$}}}{\textbf{0.05}} & & \multicolumn{1}{r@{\phantom{$\pm$}}}{0.091} & & \multirow{2}{*}{0.068} & \multirow{2}{*}{164.4} & \multirow{2}{*}{16.18} \\
 & F &  & \multicolumn{1}{r@{\phantom{$\pm$}}}{\textbf{2.98}} & & \multicolumn{1}{r@{\phantom{$\pm$}}}{0.91} & & \multicolumn{1}{r@{\phantom{$\pm$}}}{0.96} & & \multicolumn{1}{r@{\phantom{$\pm$}}}{\textbf{1.25}} & & \multicolumn{1}{r@{\phantom{$\pm$}}}{\textbf{2.59}} & & \multicolumn{1}{r@{\phantom{$\pm$}}}{\textbf{1.78}} & & \multicolumn{1}{r@{\phantom{$\pm$}}}{\textbf{2.14}} & & \multicolumn{1}{r@{\phantom{$\pm$}}}{\textbf{1.80}} & &  &  &  \\
\cmidrule(r){1-3}\cmidrule(lr){4-19}\cmidrule(l){20-22}
PET-S~\citep{Bigi2026} & E & 14.0M & \multicolumn{1}{r@{\phantom{$\pm$}}}{0.19} & & \multicolumn{1}{r@{\phantom{$\pm$}}}{0.19} & & \multicolumn{1}{r@{\phantom{$\pm$}}}{0.18} & & \multicolumn{1}{r@{\phantom{$\pm$}}}{0.11} & & \multicolumn{1}{r@{\phantom{$\pm$}}}{0.36} & & \multicolumn{1}{r@{\phantom{$\pm$}}}{0.27} & & \multicolumn{1}{r@{\phantom{$\pm$}}}{0.15} & & \multicolumn{1}{r@{\phantom{$\pm$}}}{0.207} & & \multirow{2}{*}{0.174} & \multirow{2}{*}{41.9} & \multirow{2}{*}{5.61} \\
 & F &  & \multicolumn{1}{r@{\phantom{$\pm$}}}{6.79} & & \multicolumn{1}{r@{\phantom{$\pm$}}}{2.23} & & \multicolumn{1}{r@{\phantom{$\pm$}}}{2.17} & & \multicolumn{1}{r@{\phantom{$\pm$}}}{3.58} & & \multicolumn{1}{r@{\phantom{$\pm$}}}{8.78} & & \multicolumn{1}{r@{\phantom{$\pm$}}}{6.11} & & \multicolumn{1}{r@{\phantom{$\pm$}}}{6.71} & & \multicolumn{1}{r@{\phantom{$\pm$}}}{5.20} & &  &  &  \\
PET-L~\citep{Bigi2026} & E & 188.9M & \multicolumn{1}{r@{\phantom{$\pm$}}}{\textbf{0.09}} & & \multicolumn{1}{r@{\phantom{$\pm$}}}{0.10} & & \multicolumn{1}{r@{\phantom{$\pm$}}}{0.12} & & \multicolumn{1}{r@{\phantom{$\pm$}}}{0.05} & & \multicolumn{1}{r@{\phantom{$\pm$}}}{0.17} & & \multicolumn{1}{r@{\phantom{$\pm$}}}{0.13} & & \multicolumn{1}{r@{\phantom{$\pm$}}}{0.08} & & \multicolumn{1}{r@{\phantom{$\pm$}}}{0.106} & & \multirow{2}{*}{0.086} & \multirow{2}{*}{215.7} & \multirow{2}{*}{15.40} \\
 & F &  & \multicolumn{1}{r@{\phantom{$\pm$}}}{3.53} & & \multicolumn{1}{r@{\phantom{$\pm$}}}{1.00} & & \multicolumn{1}{r@{\phantom{$\pm$}}}{1.20} & & \multicolumn{1}{r@{\phantom{$\pm$}}}{1.55} & & \multicolumn{1}{r@{\phantom{$\pm$}}}{4.37} & & \multicolumn{1}{r@{\phantom{$\pm$}}}{3.05} & & \multicolumn{1}{r@{\phantom{$\pm$}}}{2.91} & & \multicolumn{1}{r@{\phantom{$\pm$}}}{2.52} & &  &  &  \\
\cmidrule(r){1-3}\cmidrule(lr){4-19}\cmidrule(l){20-22}
FlashCart-69k & E & 69k & 1.39 & 0.07 & 0.67 & 0.09 & 0.72 & 0.04 & 0.63 & 0.03 & 1.42 & 0.26 & 0.71 & 0.05 & 0.67 & 0.02 & 0.89 & 0.07 & \multirow{2}{*}{0.669} & \multirow{2}{*}{2.4} & \multirow{2}{*}{0.80} \\
 & F &  & 26.45 & 0.03 & 10.38 & 0.02 & 10.05 & 0.13 & 15.26 & 0.13 & 22.87 & 0.41 & 14.49 & 0.41 & 24.94 & 0.18 & 17.78 & 0.05 &  &  &  \\
FlashCart-0.5M & E & 464k & 0.30 & 0.06 & 0.21 & 0.07 & 0.20 & 0.03 & 0.17 & 0.02 & 0.44 & 0.12 & 0.24 & 0.01 & 0.21 & 0.04 & 0.25 & 0.05 & \multirow{2}{*}{0.195} & \multirow{2}{*}{5.3} & \multirow{2}{*}{1.51} \\
 & F &  & 7.71 & 0.19 & 2.62 & 0.11 & 2.82 & 0.08 & 4.54 & 0.13 & 7.33 & 0.12 & 4.63 & 0.18 & 7.66 & 0.20 & 5.33 & 0.14 &  &  &  \\
FlashCart-1.1M & E & 1.1M & 0.16 & 0.00 & 0.11 & 0.01 & 0.12 & 0.00 & 0.09 & 0.00 & 0.18 & 0.01 & 0.14 & 0.01 & 0.12 & 0.01 & 0.13 & 0.01 & \multirow{2}{*}{0.115} & \multirow{2}{*}{8.4} & \multirow{2}{*}{2.62} \\
 & F &  & 5.41 & 0.04 & 1.64 & 0.02 & 1.77 & 0.02 & 2.95 & 0.03 & 4.82 & 0.09 & 3.08 & 0.08 & 5.26 & 0.03 & 3.56 & 0.04 &  &  &  \\
FlashCart-3.0M & E & 3.0M & 0.11 & 0.00 & 0.09 & 0.00 & 0.08 & 0.00 & 0.06 & 0.00 & 0.12 & 0.01 & 0.10 & 0.02 & 0.09 & 0.00 & 0.09 & 0.00 & \multirow{2}{*}{0.082} & \multirow{2}{*}{13.4} & \multirow{2}{*}{3.47} \\
 & F &  & 3.97 & 0.02 & 1.10 & 0.01 & 1.16 & 0.01 & 2.00 & 0.01 & 3.53 & 0.11 & 2.33 & 0.11 & 3.67 & 0.03 & 2.54 & 0.03 &  &  &  \\
FlashCart-5.6M & E & 5.6M & \textbf{0.09} & \textbf{0.00} & \textbf{0.07} & \textbf{0.01} & \textbf{0.06} & \textbf{0.00} & \textbf{0.04} & \textbf{0.00} & \textbf{0.08} & \textbf{0.00} & \textbf{0.08} & \textbf{0.00} & 0.07 & 0.00 & \textbf{0.07} & \textbf{0.00} & \multirow{2}{*}{\textbf{0.063}} & \multirow{2}{*}{21.5} & \multirow{2}{*}{5.24} \\
 & F &  & 3.26 & 0.03 & \textbf{0.87} & \textbf{0.02} & \textbf{0.89} & \textbf{0.01} & 1.53 & 0.02 & 2.73 & 0.06 & \textbf{1.78} & \textbf{0.07} & 2.86 & 0.06 & 1.99 & 0.04 &  &  &  \\
\bottomrule
\end{tabular}
}
\par\smallskip
{\raggedright\footnotesize $^\dagger$ Runtimes for eSEN are estimated from published results for an NVIDIA A100 GPU (80 GB)~\citep{Fu2025}.\par}
\end{table}

\begin{figure}[tbp]
\centering
\includegraphics[width=\textwidth]{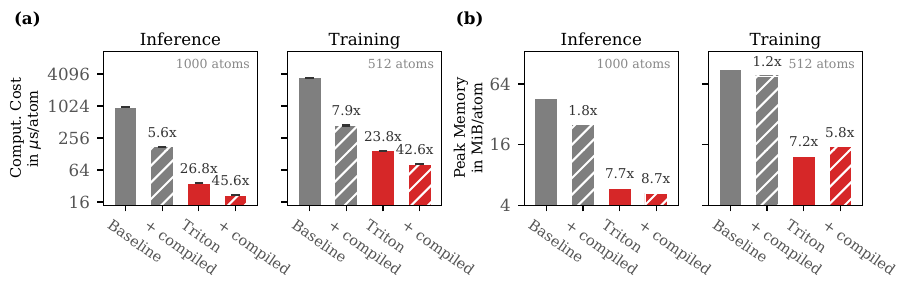}
\caption{\textbf{Runtime and memory with generated kernels and model compilation.} Runtime~(a) and peak GPU memory (b) per atom are compared for the same $5.6$M-parameter FlashCart model during inference (left) and training (right). Baseline denotes the PyTorch implementation of tensor products and edge convolutions, while Triton denotes the generated kernels. We evaluate both implementations with and without model compilation, using the fastest measured compilation mode. The factors above the bars show the runtime and memory improvements relative to the uncompiled PyTorch baseline. Inference evaluates energies and forces on $1000$ diamond atoms. Training evaluates the energy and force loss and its parameter gradients on $512$ atoms, excluding optimizer updates. Runtime error bars show standard deviations over repeated evaluations.}
\label{fig:backend}
\end{figure}

\textbf{Model accuracy and computational cost.} Table~\ref{tab:pareto-data} provides the subset errors underlying the accuracy--efficiency comparison in Figure~\ref{fig:pareto}. The largest FlashCart model uses $5.6$M parameters compared with DPA4-Plus' $8.8$M and evaluates energies and forces $7.7\times$ faster with $3.1\times$ less peak GPU memory. It achieves a lower mean energy MAE and a force MAE of $1.99\pm0.04$~meV/\AA\ compared with $1.80$~meV/\AA, as well as a lower combined normalized error. FlashCart has lower force errors than DPA4-Plus on DES370K monomers and dimers and the same reported force MAE on water. Compared with the similarly sized $5.1$M-parameter DPA4-Air, it achieves lower energy and force MAEs on all seven subsets.

Without model compilation, replacing PyTorch tensor products and edge convolutions with the generated Triton kernels makes the same model $26.8\times$ faster for inference and $23.8\times$ faster in the training benchmark (Figure~\ref{fig:backend}). Compiling the model with Triton kernels further reduces runtime but increases peak GPU memory during training. To locate this memory increase, we compare memory allocation before and during loss differentiation on $512$ atoms, with CUDA graphs disabled in the compiled version. During training we first evaluate the energy and differentiate it with respect to atomic positions to obtain forces. After the energy and force errors have been combined into the loss, GPU allocation has increased by $5.21$~GiB with compilation and $5.72$~GiB without compilation. The loss is then differentiated with respect to model parameters. During this step, allocation peaks a further $2.26$~GiB above the preceding value with compilation, compared with $0.17$~GiB without it. Thus, the additional peak memory arises during parameter-gradient computation when the model is compiled, with the same generated Triton kernels used in both cases.

\begin{table}[tbp]
\centering
\caption{\textbf{Energy and force errors on the full MAD test set.} Each energy/force pair gives the energy MAE in meV/atom followed by the force MAE in meV/\AA. FlashCart results show the mean and standard deviation across three models trained with different random seeds on a fixed data split. The overall MAEs are computed over the full test set. The lowest energy and force MAEs for each dataset are shown in bold.}
\label{tab:mad-test}
\begin{tabular}{l@{\hspace{1.5em}}w{r}{5.2em}@{\,$/$\,}w{l}{5.2em}@{\hspace{1.5em}}w{r}{2.5em}@{\,$/$\,}w{l}{2.5em}}
\toprule
Subset & \multicolumn{2}{@{}c@{\hspace{1.5em}}}{FlashCart} & \multicolumn{2}{@{}c}{PET-MAD} \\
\midrule
MC2D & \textbf{12.56}$\bm{\pm}$\textbf{0.30} & \textbf{57.2}$\bm{\pm}$\textbf{1.2} & 15.54 & 72.7 \\
MC3D & \textbf{7.02}$\bm{\pm}$\textbf{0.07} & \textbf{35.1}$\bm{\pm}$\textbf{0.4} & 9.74 & 46.3 \\
MC3D Clusters & \textbf{28.41}$\bm{\pm}$\textbf{0.54} & \textbf{127.6}$\bm{\pm}$\textbf{1.0} & 43.06 & 148.3 \\
MC3D Random & 68.09$\pm$0.71 & 316.2$\pm$2.0 & \textbf{63.87} & \textbf{249.6} \\
MC3D Rattled & \textbf{10.04}$\bm{\pm}$\textbf{0.20} & 142.2$\pm$1.1 & 11.73 & \textbf{140.3} \\
MC3D Surfaces & \textbf{10.58}$\bm{\pm}$\textbf{0.07} & \textbf{88.7}$\bm{\pm}$\textbf{1.0} & 13.61 & 102.6 \\
SHIFTML Molecular Crystals & \textbf{0.90}$\bm{\pm}$\textbf{0.01} & \textbf{21.4}$\bm{\pm}$\textbf{0.2} & 2.53 & 33.8 \\
SHIFTML Molecular Fragments & \textbf{1.18}$\bm{\pm}$\textbf{0.05} & \textbf{20.8}$\bm{\pm}$\textbf{0.4} & 6.34 & 57.1 \\
\midrule
Overall & \textbf{11.38}$\bm{\pm}$\textbf{0.13} & \textbf{63.9}$\bm{\pm}$\textbf{0.5} & 14.72 & 72.9 \\
\bottomrule
\end{tabular}
\end{table}

\textbf{Accuracy across materials and molecules.} Table~\ref{tab:mad-test} compares the energy and force errors of FlashCart and PET-MAD, covering all $9{,}546$ MAD test structures and complementing Table~\ref{tab:mad}.

\textbf{Molecular dynamics of bulk liquid water.} To demonstrate the applicability of FlashCart models to atomistic simulations, we study bulk liquid water using molecular dynamics. We use the five models in Figure~\ref{fig:pareto} to simulate $256$ water molecules in LAMMPS~\citep{Thompson2022} with Kokkos acceleration~\citep{Johansson2025} and the ML-IAP-Kokkos interface~\citep{Smith2025}. All simulations begin with energy minimization and use periodic boundary conditions and a $0.5$~fs integration timestep. For the NVE runs, we first equilibrate each system for $50$~ps at $300$~K in the NVT ensemble using a Nos\'e--Hoover chain thermostat~\citep{Martyna1992} with a time constant of $0.2$~ps, followed by $1.2$~ns of NVE dynamics. The NPT simulations are performed over $1.2$~ns after energy minimization at temperatures from $273.15$ to $373.15$~K and a pressure of $1$~bar, using a Nos\'e--Hoover chain thermostat and an isotropic Martyna--Tobias--Klein barostat with a time constant of $1$~ps~\citep{Martyna1994,Shinoda2004}.

\begin{figure}[tbp]
\centering
\includegraphics[width=\textwidth]{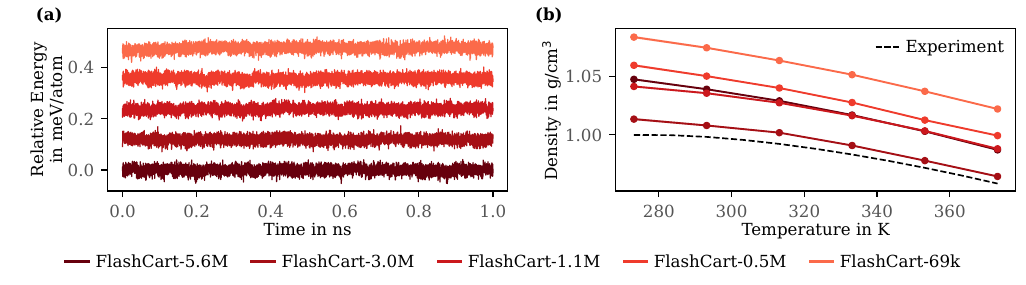}
\caption{\textbf{Energy conservation and density of bulk liquid water.} (a)~Total energy per atom during NVE dynamics, shown relative to its mean. We add a constant vertical offset to keep the curves distinguishable. (b)~Density from NPT simulations at $1$~bar, with shaded bands showing the standard error estimated by block averaging over $50$~ps intervals. The bands are narrower than the plotted lines and are therefore not visible. The dashed curve shows experimental densities at $1$~bar~\citep{Lide2005}. All results use one trained model per configuration and the final $1$~ns of each trajectory.}
\label{fig:water-md}
\end{figure}

We analyze the final $1$~ns of each trajectory to assess energy conservation in the NVE simulations and determine equilibrium densities in the NPT simulations. All models show energy drifts between $-0.002$ and $0.008$~meV/atom/ns, with energy fluctuations around the mean having standard deviations of $0.012$--$0.013$~meV/atom and maximum absolute deviations below $0.056$~meV/atom (Figure~\ref{fig:water-md}(a)). In the NPT simulations, the models predict water densities that decrease with temperature and remain above the experimental values (Figure~\ref{fig:water-md}(b)), consistent with the overestimation reported for MACE-OFF and some ICTP variants trained at the same electronic-structure level~\citep{Kovacs2025,Zaverkin2026}. Similar overestimation is also reported for MACE-POLAR-1, trained at a related electronic-structure level~\citep{Batatia2026}.

\end{document}